\documentclass[11pt,draftcls,onecolumn,journal]{IEEEtran}
\usepackage{mathtools}  

\usepackage{amsmath}
\usepackage[ pdftex]{hyperref}
\usepackage[capitalise]{cleveref}
\usepackage{caption}

\usepackage{subcaption}
\usepackage{graphicx}

\usepackage[dvipsnames]{xcolor}

\definecolor{white1}{rgb}{1.00,1.00,1.00}
\definecolor{red1}{rgb}{1.00,0.00,0.00}

\definecolor{myorange}{rgb}{0.95,0.45,0.2}      
\definecolor{mypink}{rgb}{0.90,0.1,0.9}         
\definecolor{mybluesea}{rgb}{0.0,0.85,0.85}     
\definecolor{mycobalt}{rgb}{0.35,0.35,0.9}      
\definecolor{myyellow}{rgb}{0.95,0.95,0.2}      
\definecolor{mygray}{rgb}{0.6,0.6,0.6}          

\definecolor{frontal}{rgb}{0.93,0.27,0.21}
\definecolor{colorFrontal}{rgb}{0.93,0.27,0.21}
\definecolor{lateral}{rgb}{0.68,0.66,0.80}
\definecolor{colorLateral}{rgb}{0.68,0.66,0.80}
\definecolor{horizontal}{rgb}{0.94,0.63,0.37}
\definecolor{colorHorizontal}{rgb}{0.94,0.63,0.37}

\definecolor{frontal11}{RGB}{255, 163, 163}
\definecolor{frontal12}{RGB}{255, 143, 143}
\definecolor{frontal13}{RGB}{255, 122, 122}
\definecolor{frontal14}{RGB}{255, 102, 102}
\definecolor{frontal21}{RGB}{255, 82, 82}
\definecolor{frontal22}{RGB}{255, 61, 61}
\definecolor{frontal23}{RGB}{255, 41, 41}
\definecolor{frontal24}{RGB}{255, 20, 20}
\definecolor{frontal31}{RGB}{255, 0, 0}
\definecolor{frontal32}{RGB}{235, 0, 0}
\definecolor{frontal33}{RGB}{214, 0, 0}
\definecolor{frontal34}{RGB}{194, 0, 0}
\definecolor{frontal41}{RGB}{174, 0, 0}
\definecolor{frontal42}{RGB}{154, 0, 0}
\definecolor{frontal43}{RGB}{134, 0, 0}
\definecolor{frontal44}{RGB}{114, 0, 0}

\definecolor{lateral11}{RGB}{213, 210, 229}
\definecolor{lateral12}{RGB}{200, 197, 221}
\definecolor{lateral13}{RGB}{188, 184, 214}
\definecolor{lateral14}{RGB}{176, 171, 206}
\definecolor{lateral21}{RGB}{164, 159, 198}
\definecolor{lateral22}{RGB}{152, 146, 191}
\definecolor{lateral23}{RGB}{140, 133, 183}
\definecolor{lateral24}{RGB}{128, 120, 176}
\definecolor{lateral31}{RGB}{116, 107, 168}
\definecolor{lateral32}{RGB}{103, 94, 161}
\definecolor{lateral33}{RGB}{95, 87, 148}
\definecolor{lateral34}{RGB}{87, 79, 135}
\definecolor{lateral41}{RGB}{79, 72, 122}
\definecolor{lateral42}{RGB}{70, 64, 109}
\definecolor{lateral43}{RGB}{62, 57, 96}
\definecolor{lateral44}{RGB}{54, 49, 84}

\definecolor{horizontal11}{RGB}{247, 205, 171}
\definecolor{horizontal12}{RGB}{245, 194, 152}
\definecolor{horizontal13}{RGB}{244, 183, 134}
\definecolor{horizontal14}{RGB}{242, 172, 115}
\definecolor{horizontal21}{RGB}{240, 161, 96}
\definecolor{horizontal22}{RGB}{238, 150, 78}
\definecolor{horizontal23}{RGB}{237, 139, 59}
\definecolor{horizontal24}{RGB}{235, 128, 40}
\definecolor{horizontal31}{RGB}{233, 117, 22}
\definecolor{horizontal32}{RGB}{215, 108, 20}
\definecolor{horizontal33}{RGB}{196, 99, 18}
\definecolor{horizontal34}{RGB}{177, 89, 17}
\definecolor{horizontal41}{RGB}{159, 80, 15}
\definecolor{horizontal42}{RGB}{140, 70, 13}
\definecolor{horizontal43}{RGB}{121, 61, 11}
\definecolor{horizontal44}{RGB}{103, 52, 10}

\definecolor{tnode}{rgb}{0.93,0.27,0.21}
\definecolor{mnode}{rgb}{0.68,0.66,0.80}

\definecolor{unsure1}{rgb}{1.00,0.00,0.46}

\definecolor{AliceBlue}{rgb}{0.94,0.97,1.00}
\definecolor{AntiqueWhite1}{rgb}{1.00,0.94,0.86}
\definecolor{AntiqueWhite2}{rgb}{0.93,0.87,0.80}
\definecolor{AntiqueWhite3}{rgb}{0.80,0.75,0.69}
\definecolor{AntiqueWhite4}{rgb}{0.55,0.51,0.47}
\definecolor{AntiqueWhite}{rgb}{0.98,0.92,0.84}
\definecolor{BlanchedAlmond}{rgb}{1.00,0.92,0.80}
\definecolor{BlueViolet}{rgb}{0.54,0.17,0.89}
\definecolor{CadetBlue1}{rgb}{0.60,0.96,1.00}
\definecolor{CadetBlue2}{rgb}{0.56,0.90,0.93}
\definecolor{CadetBlue3}{rgb}{0.48,0.77,0.80}
\definecolor{CadetBlue4}{rgb}{0.33,0.53,0.55}
\definecolor{CadetBlue}{rgb}{0.37,0.62,0.63}
\definecolor{CornflowerBlue}{rgb}{0.39,0.58,0.93}
\definecolor{DarkBlue}{rgb}{0.00,0.00,0.55}
\definecolor{DarkCyan}{rgb}{0.00,0.55,0.55}
\definecolor{DarkGoldenrod1}{rgb}{1.00,0.73,0.06}
\definecolor{DarkGoldenrod2}{rgb}{0.93,0.68,0.05}
\definecolor{DarkGoldenrod3}{rgb}{0.80,0.58,0.05}
\definecolor{DarkGoldenrod4}{rgb}{0.55,0.40,0.03}
\definecolor{DarkGoldenrod}{rgb}{0.72,0.53,0.04}
\definecolor{DarkGray}{rgb}{0.66,0.66,0.66}
\definecolor{DarkGreen}{rgb}{0.00,0.39,0.00}
\definecolor{DarkGrey}{rgb}{0.66,0.66,0.66}
\definecolor{DarkKhaki}{rgb}{0.74,0.72,0.42}
\definecolor{DarkMagenta}{rgb}{0.55,0.00,0.55}
\definecolor{DarkOliveGreen1}{rgb}{0.79,1.00,0.44}
\definecolor{DarkOliveGreen2}{rgb}{0.74,0.93,0.41}
\definecolor{DarkOliveGreen3}{rgb}{0.64,0.80,0.35}
\definecolor{DarkOliveGreen4}{rgb}{0.43,0.55,0.24}
\definecolor{DarkOliveGreen}{rgb}{0.33,0.42,0.18}
\definecolor{DarkOrange1}{rgb}{1.00,0.50,0.00}
\definecolor{DarkOrange2}{rgb}{0.93,0.46,0.00}
\definecolor{DarkOrange3}{rgb}{0.80,0.40,0.00}
\definecolor{DarkOrange4}{rgb}{0.55,0.27,0.00}
\definecolor{DarkOrange}{rgb}{1.00,0.55,0.00}
\definecolor{DarkOrchid1}{rgb}{0.75,0.24,1.00}
\definecolor{DarkOrchid2}{rgb}{0.70,0.23,0.93}
\definecolor{DarkOrchid3}{rgb}{0.60,0.20,0.80}
\definecolor{DarkOrchid4}{rgb}{0.41,0.13,0.55}
\definecolor{DarkOrchid}{rgb}{0.60,0.20,0.80}
\definecolor{DarkRed}{rgb}{0.55,0.00,0.00}
\definecolor{DarkSalmon}{rgb}{0.91,0.59,0.48}
\definecolor{DarkSeaGreen1}{rgb}{0.76,1.00,0.76}
\definecolor{DarkSeaGreen2}{rgb}{0.71,0.93,0.71}
\definecolor{DarkSeaGreen3}{rgb}{0.61,0.80,0.61}
\definecolor{DarkSeaGreen4}{rgb}{0.41,0.55,0.41}
\definecolor{DarkSeaGreen}{rgb}{0.56,0.74,0.56}
\definecolor{DarkSlateBlue}{rgb}{0.28,0.24,0.55}
\definecolor{DarkSlateGray1}{rgb}{0.59,1.00,1.00}
\definecolor{DarkSlateGray2}{rgb}{0.55,0.93,0.93}
\definecolor{DarkSlateGray3}{rgb}{0.47,0.80,0.80}
\definecolor{DarkSlateGray4}{rgb}{0.32,0.55,0.55}
\definecolor{DarkSlateGray}{rgb}{0.18,0.31,0.31}
\definecolor{DarkSlateGrey}{rgb}{0.18,0.31,0.31}
\definecolor{DarkTurquoise}{rgb}{0.00,0.81,0.82}
\definecolor{DarkViolet}{rgb}{0.58,0.00,0.83}
\definecolor{DeepPink1}{rgb}{1.00,0.08,0.58}
\definecolor{DeepPink2}{rgb}{0.93,0.07,0.54}
\definecolor{DeepPink3}{rgb}{0.80,0.06,0.46}
\definecolor{DeepPink4}{rgb}{0.55,0.04,0.31}
\definecolor{DeepPink}{rgb}{1.00,0.08,0.58}
\definecolor{DeepSkyBlue1}{rgb}{0.00,0.75,1.00}
\definecolor{DeepSkyBlue2}{rgb}{0.00,0.70,0.93}
\definecolor{DeepSkyBlue3}{rgb}{0.00,0.60,0.80}
\definecolor{DeepSkyBlue4}{rgb}{0.00,0.41,0.55}
\definecolor{DeepSkyBlue}{rgb}{0.00,0.75,1.00}
\definecolor{DimGray}{rgb}{0.41,0.41,0.41}
\definecolor{DimGrey}{rgb}{0.41,0.41,0.41}
\definecolor{DodgerBlue1}{rgb}{0.12,0.56,1.00}
\definecolor{DodgerBlue2}{rgb}{0.11,0.53,0.93}
\definecolor{DodgerBlue3}{rgb}{0.09,0.45,0.80}
\definecolor{DodgerBlue4}{rgb}{0.06,0.31,0.55}
\definecolor{DodgerBlue}{rgb}{0.12,0.56,1.00}
\definecolor{FloralWhite}{rgb}{1.00,0.98,0.94}
\definecolor{ForestGreen}{rgb}{0.13,0.55,0.13}
\definecolor{GhostWhite}{rgb}{0.97,0.97,1.00}
\definecolor{GreenYellow}{rgb}{0.68,1.00,0.18}
\definecolor{HotPink1}{rgb}{1.00,0.43,0.71}
\definecolor{HotPink2}{rgb}{0.93,0.42,0.65}
\definecolor{HotPink3}{rgb}{0.80,0.38,0.56}
\definecolor{HotPink4}{rgb}{0.55,0.23,0.38}
\definecolor{HotPink}{rgb}{1.00,0.41,0.71}
\definecolor{IndianRed1}{rgb}{1.00,0.42,0.42}
\definecolor{IndianRed2}{rgb}{0.93,0.39,0.39}
\definecolor{IndianRed3}{rgb}{0.80,0.33,0.33}
\definecolor{IndianRed4}{rgb}{0.55,0.23,0.23}
\definecolor{IndianRed}{rgb}{0.80,0.36,0.36}
\definecolor{LavenderBlush1}{rgb}{1.00,0.94,0.96}
\definecolor{LavenderBlush2}{rgb}{0.93,0.88,0.90}
\definecolor{LavenderBlush3}{rgb}{0.80,0.76,0.77}
\definecolor{LavenderBlush4}{rgb}{0.55,0.51,0.53}
\definecolor{LavenderBlush}{rgb}{1.00,0.94,0.96}
\definecolor{LawnGreen}{rgb}{0.49,0.99,0.00}
\definecolor{LemonChiffon1}{rgb}{1.00,0.98,0.80}
\definecolor{LemonChiffon2}{rgb}{0.93,0.91,0.75}
\definecolor{LemonChiffon3}{rgb}{0.80,0.79,0.65}
\definecolor{LemonChiffon4}{rgb}{0.55,0.54,0.44}
\definecolor{LemonChiffon}{rgb}{1.00,0.98,0.80}
\definecolor{LightBlue1}{rgb}{0.75,0.94,1.00}
\definecolor{LightBlue2}{rgb}{0.70,0.87,0.93}
\definecolor{LightBlue3}{rgb}{0.60,0.75,0.80}
\definecolor{LightBlue4}{rgb}{0.41,0.51,0.55}
\definecolor{LightBlue}{rgb}{0.68,0.85,0.90}
\definecolor{LightCoral}{rgb}{0.94,0.50,0.50}
\definecolor{LightCyan1}{rgb}{0.88,1.00,1.00}
\definecolor{LightCyan2}{rgb}{0.82,0.93,0.93}
\definecolor{LightCyan3}{rgb}{0.71,0.80,0.80}
\definecolor{LightCyan4}{rgb}{0.48,0.55,0.55}
\definecolor{LightCyan}{rgb}{0.88,1.00,1.00}
\definecolor{LightGoldenrod1}{rgb}{1.00,0.93,0.55}
\definecolor{LightGoldenrod2}{rgb}{0.93,0.86,0.51}
\definecolor{LightGoldenrod3}{rgb}{0.80,0.75,0.44}
\definecolor{LightGoldenrod4}{rgb}{0.55,0.51,0.30}
\definecolor{LightGoldenrodYellow}{rgb}{0.98,0.98,0.82}
\definecolor{LightGoldenrod}{rgb}{0.93,0.87,0.51}
\definecolor{LightGray}{rgb}{0.83,0.83,0.83}
\definecolor{LightGreen}{rgb}{0.56,0.93,0.56}
\definecolor{LightGrey}{rgb}{0.83,0.83,0.83}
\definecolor{LightPink1}{rgb}{1.00,0.68,0.73}
\definecolor{LightPink2}{rgb}{0.93,0.64,0.68}
\definecolor{LightPink3}{rgb}{0.80,0.55,0.58}
\definecolor{LightPink4}{rgb}{0.55,0.37,0.40}
\definecolor{LightPink}{rgb}{1.00,0.71,0.76}
\definecolor{LightSalmon1}{rgb}{1.00,0.63,0.48}
\definecolor{LightSalmon2}{rgb}{0.93,0.58,0.45}
\definecolor{LightSalmon3}{rgb}{0.80,0.51,0.38}
\definecolor{LightSalmon4}{rgb}{0.55,0.34,0.26}
\definecolor{LightSalmon}{rgb}{1.00,0.63,0.48}
\definecolor{LightSeaGreen}{rgb}{0.13,0.70,0.67}
\definecolor{LightSkyBlue1}{rgb}{0.69,0.89,1.00}
\definecolor{LightSkyBlue2}{rgb}{0.64,0.83,0.93}
\definecolor{LightSkyBlue3}{rgb}{0.55,0.71,0.80}
\definecolor{LightSkyBlue4}{rgb}{0.38,0.48,0.55}
\definecolor{LightSkyBlue}{rgb}{0.53,0.81,0.98}
\definecolor{LightSlateBlue}{rgb}{0.52,0.44,1.00}
\definecolor{LightSlateGray}{rgb}{0.47,0.53,0.60}
\definecolor{LightSlateGrey}{rgb}{0.47,0.53,0.60}
\definecolor{LightSteelBlue1}{rgb}{0.79,0.88,1.00}
\definecolor{LightSteelBlue2}{rgb}{0.74,0.82,0.93}
\definecolor{LightSteelBlue3}{rgb}{0.64,0.71,0.80}
\definecolor{LightSteelBlue4}{rgb}{0.43,0.48,0.55}
\definecolor{LightSteelBlue}{rgb}{0.69,0.77,0.87}
\definecolor{LightYellow1}{rgb}{1.00,1.00,0.88}
\definecolor{LightYellow2}{rgb}{0.93,0.93,0.82}
\definecolor{LightYellow3}{rgb}{0.80,0.80,0.71}
\definecolor{LightYellow4}{rgb}{0.55,0.55,0.48}
\definecolor{LightYellow}{rgb}{1.00,1.00,0.88}
\definecolor{LimeGreen}{rgb}{0.20,0.80,0.20}
\definecolor{MediumAquamarine}{rgb}{0.40,0.80,0.67}
\definecolor{MediumBlue}{rgb}{0.00,0.00,0.80}
\definecolor{MediumOrchid1}{rgb}{0.88,0.40,1.00}
\definecolor{MediumOrchid2}{rgb}{0.82,0.37,0.93}
\definecolor{MediumOrchid3}{rgb}{0.71,0.32,0.80}
\definecolor{MediumOrchid4}{rgb}{0.48,0.22,0.55}
\definecolor{MediumOrchid}{rgb}{0.73,0.33,0.83}
\definecolor{MediumPurple1}{rgb}{0.67,0.51,1.00}
\definecolor{MediumPurple2}{rgb}{0.62,0.47,0.93}
\definecolor{MediumPurple3}{rgb}{0.54,0.41,0.80}
\definecolor{MediumPurple4}{rgb}{0.36,0.28,0.55}
\definecolor{MediumPurple}{rgb}{0.58,0.44,0.86}
\definecolor{MediumSeaGreen}{rgb}{0.24,0.70,0.44}
\definecolor{MediumSlateBlue}{rgb}{0.48,0.41,0.93}
\definecolor{MediumSpringGreen}{rgb}{0.00,0.98,0.60}
\definecolor{MediumTurquoise}{rgb}{0.28,0.82,0.80}
\definecolor{MediumVioletRed}{rgb}{0.78,0.08,0.52}
\definecolor{MidnightBlue}{rgb}{0.10,0.10,0.44}
\definecolor{MintCream}{rgb}{0.96,1.00,0.98}
\definecolor{MistyRose1}{rgb}{1.00,0.89,0.88}
\definecolor{MistyRose2}{rgb}{0.93,0.84,0.82}
\definecolor{MistyRose3}{rgb}{0.80,0.72,0.71}
\definecolor{MistyRose4}{rgb}{0.55,0.49,0.48}
\definecolor{MistyRose}{rgb}{1.00,0.89,0.88}
\definecolor{NavajoWhite1}{rgb}{1.00,0.87,0.68}
\definecolor{NavajoWhite2}{rgb}{0.93,0.81,0.63}
\definecolor{NavajoWhite3}{rgb}{0.80,0.70,0.55}
\definecolor{NavajoWhite4}{rgb}{0.55,0.47,0.37}
\definecolor{NavajoWhite}{rgb}{1.00,0.87,0.68}
\definecolor{NavyBlue}{rgb}{0.00,0.00,0.50}
\definecolor{OldLace}{rgb}{0.99,0.96,0.90}
\definecolor{OliveDrab1}{rgb}{0.75,1.00,0.24}
\definecolor{OliveDrab2}{rgb}{0.70,0.93,0.23}
\definecolor{OliveDrab3}{rgb}{0.60,0.80,0.20}
\definecolor{OliveDrab4}{rgb}{0.41,0.55,0.13}
\definecolor{OliveDrab}{rgb}{0.42,0.56,0.14}
\definecolor{OrangeRed1}{rgb}{1.00,0.27,0.00}
\definecolor{OrangeRed2}{rgb}{0.93,0.25,0.00}
\definecolor{OrangeRed3}{rgb}{0.80,0.22,0.00}
\definecolor{OrangeRed4}{rgb}{0.55,0.15,0.00}
\definecolor{OrangeRed}{rgb}{1.00,0.27,0.00}
\definecolor{PaleGoldenrod}{rgb}{0.93,0.91,0.67}
\definecolor{PaleGreen1}{rgb}{0.60,1.00,0.60}
\definecolor{PaleGreen2}{rgb}{0.56,0.93,0.56}
\definecolor{PaleGreen3}{rgb}{0.49,0.80,0.49}
\definecolor{PaleGreen4}{rgb}{0.33,0.55,0.33}
\definecolor{PaleGreen}{rgb}{0.60,0.98,0.60}
\definecolor{PaleTurquoise1}{rgb}{0.73,1.00,1.00}
\definecolor{PaleTurquoise2}{rgb}{0.68,0.93,0.93}
\definecolor{PaleTurquoise3}{rgb}{0.59,0.80,0.80}
\definecolor{PaleTurquoise4}{rgb}{0.40,0.55,0.55}
\definecolor{PaleTurquoise}{rgb}{0.69,0.93,0.93}
\definecolor{PaleVioletRed1}{rgb}{1.00,0.51,0.67}
\definecolor{PaleVioletRed2}{rgb}{0.93,0.47,0.62}
\definecolor{PaleVioletRed3}{rgb}{0.80,0.41,0.54}
\definecolor{PaleVioletRed4}{rgb}{0.55,0.28,0.36}
\definecolor{PaleVioletRed}{rgb}{0.86,0.44,0.58}
\definecolor{PapayaWhip}{rgb}{1.00,0.94,0.84}
\definecolor{PeachPuff1}{rgb}{1.00,0.85,0.73}
\definecolor{PeachPuff2}{rgb}{0.93,0.80,0.68}
\definecolor{PeachPuff3}{rgb}{0.80,0.69,0.58}
\definecolor{PeachPuff4}{rgb}{0.55,0.47,0.40}
\definecolor{PeachPuff}{rgb}{1.00,0.85,0.73}
\definecolor{PowderBlue}{rgb}{0.69,0.88,0.90}
\definecolor{RosyBrown1}{rgb}{1.00,0.76,0.76}
\definecolor{RosyBrown2}{rgb}{0.93,0.71,0.71}
\definecolor{RosyBrown3}{rgb}{0.80,0.61,0.61}
\definecolor{RosyBrown4}{rgb}{0.55,0.41,0.41}
\definecolor{RosyBrown}{rgb}{0.74,0.56,0.56}
\definecolor{RoyalBlue1}{rgb}{0.28,0.46,1.00}
\definecolor{RoyalBlue2}{rgb}{0.26,0.43,0.93}
\definecolor{RoyalBlue3}{rgb}{0.23,0.37,0.80}
\definecolor{RoyalBlue4}{rgb}{0.15,0.25,0.55}
\definecolor{RoyalBlue}{rgb}{0.25,0.41,0.88}
\definecolor{SaddleBrown}{rgb}{0.55,0.27,0.07}
\definecolor{SandyBrown}{rgb}{0.96,0.64,0.38}
\definecolor{SeaGreen1}{rgb}{0.33,1.00,0.62}
\definecolor{SeaGreen2}{rgb}{0.31,0.93,0.58}
\definecolor{SeaGreen3}{rgb}{0.26,0.80,0.50}
\definecolor{SeaGreen4}{rgb}{0.18,0.55,0.34}
\definecolor{SeaGreen}{rgb}{0.18,0.55,0.34}
\definecolor{SkyBlue1}{rgb}{0.53,0.81,1.00}
\definecolor{SkyBlue2}{rgb}{0.49,0.75,0.93}
\definecolor{SkyBlue3}{rgb}{0.42,0.65,0.80}
\definecolor{SkyBlue4}{rgb}{0.29,0.44,0.55}
\definecolor{SkyBlue}{rgb}{0.53,0.81,0.92}
\definecolor{SlateBlue1}{rgb}{0.51,0.44,1.00}
\definecolor{SlateBlue2}{rgb}{0.48,0.40,0.93}
\definecolor{SlateBlue3}{rgb}{0.41,0.35,0.80}
\definecolor{SlateBlue4}{rgb}{0.28,0.24,0.55}
\definecolor{SlateBlue}{rgb}{0.42,0.35,0.80}
\definecolor{SlateGray1}{rgb}{0.78,0.89,1.00}
\definecolor{SlateGray2}{rgb}{0.73,0.83,0.93}
\definecolor{SlateGray3}{rgb}{0.62,0.71,0.80}
\definecolor{SlateGray4}{rgb}{0.42,0.48,0.55}
\definecolor{SlateGray}{rgb}{0.44,0.50,0.56}
\definecolor{SlateGrey}{rgb}{0.44,0.50,0.56}
\definecolor{SpringGreen1}{rgb}{0.00,1.00,0.50}
\definecolor{SpringGreen2}{rgb}{0.00,0.93,0.46}
\definecolor{SpringGreen3}{rgb}{0.00,0.80,0.40}
\definecolor{SpringGreen4}{rgb}{0.00,0.55,0.27}
\definecolor{SpringGreen}{rgb}{0.00,1.00,0.50}
\definecolor{SteelBlue1}{rgb}{0.39,0.72,1.00}
\definecolor{SteelBlue2}{rgb}{0.36,0.67,0.93}
\definecolor{SteelBlue3}{rgb}{0.31,0.58,0.80}
\definecolor{SteelBlue4}{rgb}{0.21,0.39,0.55}
\definecolor{SteelBlue}{rgb}{0.27,0.51,0.71}
\definecolor{VioletRed1}{rgb}{1.00,0.24,0.59}
\definecolor{VioletRed2}{rgb}{0.93,0.23,0.55}
\definecolor{VioletRed3}{rgb}{0.80,0.20,0.47}
\definecolor{VioletRed4}{rgb}{0.55,0.13,0.32}
\definecolor{VioletRed}{rgb}{0.82,0.13,0.56}
\definecolor{WhiteSmoke}{rgb}{0.96,0.96,0.96}
\definecolor{YellowGreen}{rgb}{0.60,0.80,0.20}
\definecolor{aliceblue}{rgb}{0.94,0.97,1.00}
\definecolor{antiquewhite}{rgb}{0.98,0.92,0.84}
\definecolor{aquamarine1}{rgb}{0.50,1.00,0.83}
\definecolor{aquamarine2}{rgb}{0.46,0.93,0.78}
\definecolor{aquamarine3}{rgb}{0.40,0.80,0.67}
\definecolor{aquamarine4}{rgb}{0.27,0.55,0.45}
\definecolor{aquamarine}{rgb}{0.50,1.00,0.83}
\definecolor{azure1}{rgb}{0.94,1.00,1.00}
\definecolor{azure2}{rgb}{0.88,0.93,0.93}
\definecolor{azure3}{rgb}{0.76,0.80,0.80}
\definecolor{azure4}{rgb}{0.51,0.55,0.55}
\definecolor{azure}{rgb}{0.94,1.00,1.00}
\definecolor{beige}{rgb}{0.96,0.96,0.86}
\definecolor{bisque1}{rgb}{1.00,0.89,0.77}
\definecolor{bisque2}{rgb}{0.93,0.84,0.72}
\definecolor{bisque3}{rgb}{0.80,0.72,0.62}
\definecolor{bisque4}{rgb}{0.55,0.49,0.42}
\definecolor{bisque}{rgb}{1.00,0.89,0.77}
\definecolor{black}{rgb}{0.00,0.00,0.00}
\definecolor{blanchedalmond}{rgb}{1.00,0.92,0.80}
\definecolor{blue1}{rgb}{0.00,0.00,1.00}
\definecolor{blue2}{rgb}{0.00,0.00,0.93}
\definecolor{blue3}{rgb}{0.00,0.00,0.80}
\definecolor{blue4}{rgb}{0.00,0.00,0.55}
\definecolor{blueviolet}{rgb}{0.54,0.17,0.89}
\definecolor{blue}{rgb}{0.00,0.00,1.00}
\definecolor{brown1}{rgb}{1.00,0.25,0.25}
\definecolor{brown2}{rgb}{0.93,0.23,0.23}
\definecolor{brown3}{rgb}{0.80,0.20,0.20}
\definecolor{brown4}{rgb}{0.55,0.14,0.14}
\definecolor{brown}{rgb}{0.65,0.16,0.16}
\definecolor{burlywood1}{rgb}{1.00,0.83,0.61}
\definecolor{burlywood2}{rgb}{0.93,0.77,0.57}
\definecolor{burlywood3}{rgb}{0.80,0.67,0.49}
\definecolor{burlywood4}{rgb}{0.55,0.45,0.33}
\definecolor{burlywood}{rgb}{0.87,0.72,0.53}
\definecolor{cadetblue}{rgb}{0.37,0.62,0.63}
\definecolor{chartreuse1}{rgb}{0.50,1.00,0.00}
\definecolor{chartreuse2}{rgb}{0.46,0.93,0.00}
\definecolor{chartreuse3}{rgb}{0.40,0.80,0.00}
\definecolor{chartreuse4}{rgb}{0.27,0.55,0.00}
\definecolor{chartreuse}{rgb}{0.50,1.00,0.00}
\definecolor{chocolate1}{rgb}{1.00,0.50,0.14}
\definecolor{chocolate2}{rgb}{0.93,0.46,0.13}
\definecolor{chocolate3}{rgb}{0.80,0.40,0.11}
\definecolor{chocolate4}{rgb}{0.55,0.27,0.07}
\definecolor{chocolate}{rgb}{0.82,0.41,0.12}
\definecolor{coral1}{rgb}{1.00,0.45,0.34}
\definecolor{coral2}{rgb}{0.93,0.42,0.31}
\definecolor{coral3}{rgb}{0.80,0.36,0.27}
\definecolor{coral4}{rgb}{0.55,0.24,0.18}
\definecolor{coral}{rgb}{1.00,0.50,0.31}
\definecolor{cornflowerblue}{rgb}{0.39,0.58,0.93}
\definecolor{cornsilk1}{rgb}{1.00,0.97,0.86}
\definecolor{cornsilk2}{rgb}{0.93,0.91,0.80}
\definecolor{cornsilk3}{rgb}{0.80,0.78,0.69}
\definecolor{cornsilk4}{rgb}{0.55,0.53,0.47}
\definecolor{cornsilk}{rgb}{1.00,0.97,0.86}
\definecolor{cyan1}{rgb}{0.00,1.00,1.00}
\definecolor{cyan2}{rgb}{0.00,0.93,0.93}
\definecolor{cyan3}{rgb}{0.00,0.80,0.80}
\definecolor{cyan4}{rgb}{0.00,0.55,0.55}
\definecolor{cyan}{rgb}{0.00,1.00,1.00}
\definecolor{darkblue}{rgb}{0.00,0.00,0.55}
\definecolor{darkcyan}{rgb}{0.00,0.55,0.55}
\definecolor{darkgoldenrod}{rgb}{0.72,0.53,0.04}
\definecolor{darkgray}{rgb}{0.66,0.66,0.66}
\definecolor{darkgreen}{rgb}{0.00,0.39,0.00}
\definecolor{darkgrey}{rgb}{0.66,0.66,0.66}
\definecolor{darkkhaki}{rgb}{0.74,0.72,0.42}
\definecolor{darkmagenta}{rgb}{0.55,0.00,0.55}
\definecolor{darkolive}{rgb}{0.33,0.42,0.18}
\definecolor{darkorange}{rgb}{1.00,0.55,0.00}
\definecolor{darkorchid}{rgb}{0.60,0.20,0.80}
\definecolor{darkred}{rgb}{0.55,0.00,0.00}
\definecolor{darksalmon}{rgb}{0.91,0.59,0.48}
\definecolor{darksea}{rgb}{0.56,0.74,0.56}
\definecolor{darkslate}{rgb}{0.18,0.31,0.31}
\definecolor{darkslate}{rgb}{0.18,0.31,0.31}
\definecolor{darkslate}{rgb}{0.28,0.24,0.55}
\definecolor{darkturquoise}{rgb}{0.00,0.81,0.82}
\definecolor{darkviolet}{rgb}{0.58,0.00,0.83}
\definecolor{deeppink}{rgb}{1.00,0.08,0.58}
\definecolor{deepsky}{rgb}{0.00,0.75,1.00}
\definecolor{dimgray}{rgb}{0.41,0.41,0.41}
\definecolor{dimgrey}{rgb}{0.41,0.41,0.41}
\definecolor{dodgerblue}{rgb}{0.12,0.56,1.00}
\definecolor{firebrick1}{rgb}{1.00,0.19,0.19}
\definecolor{firebrick2}{rgb}{0.93,0.17,0.17}
\definecolor{firebrick3}{rgb}{0.80,0.15,0.15}
\definecolor{firebrick4}{rgb}{0.55,0.10,0.10}
\definecolor{firebrick}{rgb}{0.70,0.13,0.13}
\definecolor{floralwhite}{rgb}{1.00,0.98,0.94}
\definecolor{forestgreen}{rgb}{0.13,0.55,0.13}
\definecolor{gainsboro}{rgb}{0.86,0.86,0.86}
\definecolor{ghostwhite}{rgb}{0.97,0.97,1.00}
\definecolor{gold1}{rgb}{1.00,0.84,0.00}
\definecolor{gold2}{rgb}{0.93,0.79,0.00}
\definecolor{gold3}{rgb}{0.80,0.68,0.00}
\definecolor{gold4}{rgb}{0.55,0.46,0.00}
\definecolor{goldenrod1}{rgb}{1.00,0.76,0.15}
\definecolor{goldenrod2}{rgb}{0.93,0.71,0.13}
\definecolor{goldenrod3}{rgb}{0.80,0.61,0.11}
\definecolor{goldenrod4}{rgb}{0.55,0.41,0.08}
\definecolor{goldenrod}{rgb}{0.85,0.65,0.13}
\definecolor{gold}{rgb}{1.00,0.84,0.00}
\definecolor{gray0}{rgb}{0.00,0.00,0.00}
\definecolor{gray100}{rgb}{1.00,1.00,1.00}
\definecolor{gray10}{rgb}{0.10,0.10,0.10}
\definecolor{gray11}{rgb}{0.11,0.11,0.11}
\definecolor{gray12}{rgb}{0.12,0.12,0.12}
\definecolor{gray13}{rgb}{0.13,0.13,0.13}
\definecolor{gray14}{rgb}{0.14,0.14,0.14}
\definecolor{gray15}{rgb}{0.15,0.15,0.15}
\definecolor{gray16}{rgb}{0.16,0.16,0.16}
\definecolor{gray17}{rgb}{0.17,0.17,0.17}
\definecolor{gray18}{rgb}{0.18,0.18,0.18}
\definecolor{gray19}{rgb}{0.19,0.19,0.19}
\definecolor{gray1}{rgb}{0.01,0.01,0.01}
\definecolor{gray20}{rgb}{0.20,0.20,0.20}
\definecolor{gray21}{rgb}{0.21,0.21,0.21}
\definecolor{gray22}{rgb}{0.22,0.22,0.22}
\definecolor{gray23}{rgb}{0.23,0.23,0.23}
\definecolor{gray24}{rgb}{0.24,0.24,0.24}
\definecolor{gray25}{rgb}{0.25,0.25,0.25}
\definecolor{gray26}{rgb}{0.26,0.26,0.26}
\definecolor{gray27}{rgb}{0.27,0.27,0.27}
\definecolor{gray28}{rgb}{0.28,0.28,0.28}
\definecolor{gray29}{rgb}{0.29,0.29,0.29}
\definecolor{gray2}{rgb}{0.02,0.02,0.02}
\definecolor{gray30}{rgb}{0.30,0.30,0.30}
\definecolor{gray31}{rgb}{0.31,0.31,0.31}
\definecolor{gray32}{rgb}{0.32,0.32,0.32}
\definecolor{gray33}{rgb}{0.33,0.33,0.33}
\definecolor{gray34}{rgb}{0.34,0.34,0.34}
\definecolor{gray35}{rgb}{0.35,0.35,0.35}
\definecolor{gray36}{rgb}{0.36,0.36,0.36}
\definecolor{gray37}{rgb}{0.37,0.37,0.37}
\definecolor{gray38}{rgb}{0.38,0.38,0.38}
\definecolor{gray39}{rgb}{0.39,0.39,0.39}
\definecolor{gray3}{rgb}{0.03,0.03,0.03}
\definecolor{gray40}{rgb}{0.40,0.40,0.40}
\definecolor{gray41}{rgb}{0.41,0.41,0.41}
\definecolor{gray42}{rgb}{0.42,0.42,0.42}
\definecolor{gray43}{rgb}{0.43,0.43,0.43}
\definecolor{gray44}{rgb}{0.44,0.44,0.44}
\definecolor{gray45}{rgb}{0.45,0.45,0.45}
\definecolor{gray46}{rgb}{0.46,0.46,0.46}
\definecolor{gray47}{rgb}{0.47,0.47,0.47}
\definecolor{gray48}{rgb}{0.48,0.48,0.48}
\definecolor{gray49}{rgb}{0.49,0.49,0.49}
\definecolor{gray4}{rgb}{0.04,0.04,0.04}
\definecolor{gray50}{rgb}{0.50,0.50,0.50}
\definecolor{gray51}{rgb}{0.51,0.51,0.51}
\definecolor{gray52}{rgb}{0.52,0.52,0.52}
\definecolor{gray53}{rgb}{0.53,0.53,0.53}
\definecolor{gray54}{rgb}{0.54,0.54,0.54}
\definecolor{gray55}{rgb}{0.55,0.55,0.55}
\definecolor{gray56}{rgb}{0.56,0.56,0.56}
\definecolor{gray57}{rgb}{0.57,0.57,0.57}
\definecolor{gray58}{rgb}{0.58,0.58,0.58}
\definecolor{gray59}{rgb}{0.59,0.59,0.59}
\definecolor{gray5}{rgb}{0.05,0.05,0.05}
\definecolor{gray60}{rgb}{0.60,0.60,0.60}
\definecolor{gray61}{rgb}{0.61,0.61,0.61}
\definecolor{gray62}{rgb}{0.62,0.62,0.62}
\definecolor{gray63}{rgb}{0.63,0.63,0.63}
\definecolor{gray64}{rgb}{0.64,0.64,0.64}
\definecolor{gray65}{rgb}{0.65,0.65,0.65}
\definecolor{gray66}{rgb}{0.66,0.66,0.66}
\definecolor{gray67}{rgb}{0.67,0.67,0.67}
\definecolor{gray68}{rgb}{0.68,0.68,0.68}
\definecolor{gray69}{rgb}{0.69,0.69,0.69}
\definecolor{gray6}{rgb}{0.06,0.06,0.06}
\definecolor{gray70}{rgb}{0.70,0.70,0.70}
\definecolor{gray71}{rgb}{0.71,0.71,0.71}
\definecolor{gray72}{rgb}{0.72,0.72,0.72}
\definecolor{gray73}{rgb}{0.73,0.73,0.73}
\definecolor{gray74}{rgb}{0.74,0.74,0.74}
\definecolor{gray75}{rgb}{0.75,0.75,0.75}
\definecolor{gray76}{rgb}{0.76,0.76,0.76}
\definecolor{gray77}{rgb}{0.77,0.77,0.77}
\definecolor{gray78}{rgb}{0.78,0.78,0.78}
\definecolor{gray79}{rgb}{0.79,0.79,0.79}
\definecolor{gray7}{rgb}{0.07,0.07,0.07}
\definecolor{gray80}{rgb}{0.80,0.80,0.80}
\definecolor{gray81}{rgb}{0.81,0.81,0.81}
\definecolor{gray82}{rgb}{0.82,0.82,0.82}
\definecolor{gray83}{rgb}{0.83,0.83,0.83}
\definecolor{gray84}{rgb}{0.84,0.84,0.84}
\definecolor{gray85}{rgb}{0.85,0.85,0.85}
\definecolor{gray86}{rgb}{0.86,0.86,0.86}
\definecolor{gray87}{rgb}{0.87,0.87,0.87}
\definecolor{gray88}{rgb}{0.88,0.88,0.88}
\definecolor{gray89}{rgb}{0.89,0.89,0.89}
\definecolor{gray8}{rgb}{0.08,0.08,0.08}
\definecolor{gray90}{rgb}{0.90,0.90,0.90}
\definecolor{gray91}{rgb}{0.91,0.91,0.91}
\definecolor{gray92}{rgb}{0.92,0.92,0.92}
\definecolor{gray93}{rgb}{0.93,0.93,0.93}
\definecolor{gray94}{rgb}{0.94,0.94,0.94}
\definecolor{gray95}{rgb}{0.95,0.95,0.95}
\definecolor{gray96}{rgb}{0.96,0.96,0.96}
\definecolor{gray97}{rgb}{0.97,0.97,0.97}
\definecolor{gray98}{rgb}{0.98,0.98,0.98}
\definecolor{gray99}{rgb}{0.99,0.99,0.99}
\definecolor{gray9}{rgb}{0.09,0.09,0.09}
\definecolor{gray}{rgb}{0.75,0.75,0.75}
\definecolor{green1}{rgb}{0.00,1.00,0.00}
\definecolor{green2}{rgb}{0.00,0.93,0.00}
\definecolor{green3}{rgb}{0.00,0.80,0.00}
\definecolor{green4}{rgb}{0.00,0.55,0.00}
\definecolor{greenyellow}{rgb}{0.68,1.00,0.18}
\definecolor{green}{rgb}{0.00,1.00,0.00}
\definecolor{grey0}{rgb}{0.00,0.00,0.00}
\definecolor{grey100}{rgb}{1.00,1.00,1.00}
\definecolor{grey10}{rgb}{0.10,0.10,0.10}
\definecolor{grey11}{rgb}{0.11,0.11,0.11}
\definecolor{grey12}{rgb}{0.12,0.12,0.12}
\definecolor{grey13}{rgb}{0.13,0.13,0.13}
\definecolor{grey14}{rgb}{0.14,0.14,0.14}
\definecolor{grey15}{rgb}{0.15,0.15,0.15}
\definecolor{grey16}{rgb}{0.16,0.16,0.16}
\definecolor{grey17}{rgb}{0.17,0.17,0.17}
\definecolor{grey18}{rgb}{0.18,0.18,0.18}
\definecolor{grey19}{rgb}{0.19,0.19,0.19}
\definecolor{grey1}{rgb}{0.01,0.01,0.01}
\definecolor{grey20}{rgb}{0.20,0.20,0.20}
\definecolor{grey21}{rgb}{0.21,0.21,0.21}
\definecolor{grey22}{rgb}{0.22,0.22,0.22}
\definecolor{grey23}{rgb}{0.23,0.23,0.23}
\definecolor{grey24}{rgb}{0.24,0.24,0.24}
\definecolor{grey25}{rgb}{0.25,0.25,0.25}
\definecolor{grey26}{rgb}{0.26,0.26,0.26}
\definecolor{grey27}{rgb}{0.27,0.27,0.27}
\definecolor{grey28}{rgb}{0.28,0.28,0.28}
\definecolor{grey29}{rgb}{0.29,0.29,0.29}
\definecolor{grey2}{rgb}{0.02,0.02,0.02}
\definecolor{grey30}{rgb}{0.30,0.30,0.30}
\definecolor{grey31}{rgb}{0.31,0.31,0.31}
\definecolor{grey32}{rgb}{0.32,0.32,0.32}
\definecolor{grey33}{rgb}{0.33,0.33,0.33}
\definecolor{grey34}{rgb}{0.34,0.34,0.34}
\definecolor{grey35}{rgb}{0.35,0.35,0.35}
\definecolor{grey36}{rgb}{0.36,0.36,0.36}
\definecolor{grey37}{rgb}{0.37,0.37,0.37}
\definecolor{grey38}{rgb}{0.38,0.38,0.38}
\definecolor{grey39}{rgb}{0.39,0.39,0.39}
\definecolor{grey3}{rgb}{0.03,0.03,0.03}
\definecolor{grey40}{rgb}{0.40,0.40,0.40}
\definecolor{grey41}{rgb}{0.41,0.41,0.41}
\definecolor{grey42}{rgb}{0.42,0.42,0.42}
\definecolor{grey43}{rgb}{0.43,0.43,0.43}
\definecolor{grey44}{rgb}{0.44,0.44,0.44}
\definecolor{grey45}{rgb}{0.45,0.45,0.45}
\definecolor{grey46}{rgb}{0.46,0.46,0.46}
\definecolor{grey47}{rgb}{0.47,0.47,0.47}
\definecolor{grey48}{rgb}{0.48,0.48,0.48}
\definecolor{grey49}{rgb}{0.49,0.49,0.49}
\definecolor{grey4}{rgb}{0.04,0.04,0.04}
\definecolor{grey50}{rgb}{0.50,0.50,0.50}
\definecolor{grey51}{rgb}{0.51,0.51,0.51}
\definecolor{grey52}{rgb}{0.52,0.52,0.52}
\definecolor{grey53}{rgb}{0.53,0.53,0.53}
\definecolor{grey54}{rgb}{0.54,0.54,0.54}
\definecolor{grey55}{rgb}{0.55,0.55,0.55}
\definecolor{grey56}{rgb}{0.56,0.56,0.56}
\definecolor{grey57}{rgb}{0.57,0.57,0.57}
\definecolor{grey58}{rgb}{0.58,0.58,0.58}
\definecolor{grey59}{rgb}{0.59,0.59,0.59}
\definecolor{grey5}{rgb}{0.05,0.05,0.05}
\definecolor{grey60}{rgb}{0.60,0.60,0.60}
\definecolor{grey61}{rgb}{0.61,0.61,0.61}
\definecolor{grey62}{rgb}{0.62,0.62,0.62}
\definecolor{grey63}{rgb}{0.63,0.63,0.63}
\definecolor{grey64}{rgb}{0.64,0.64,0.64}
\definecolor{grey65}{rgb}{0.65,0.65,0.65}
\definecolor{grey66}{rgb}{0.66,0.66,0.66}
\definecolor{grey67}{rgb}{0.67,0.67,0.67}
\definecolor{grey68}{rgb}{0.68,0.68,0.68}
\definecolor{grey69}{rgb}{0.69,0.69,0.69}
\definecolor{grey6}{rgb}{0.06,0.06,0.06}
\definecolor{grey70}{rgb}{0.70,0.70,0.70}
\definecolor{grey71}{rgb}{0.71,0.71,0.71}
\definecolor{grey72}{rgb}{0.72,0.72,0.72}
\definecolor{grey73}{rgb}{0.73,0.73,0.73}
\definecolor{grey74}{rgb}{0.74,0.74,0.74}
\definecolor{grey75}{rgb}{0.75,0.75,0.75}
\definecolor{grey76}{rgb}{0.76,0.76,0.76}
\definecolor{grey77}{rgb}{0.77,0.77,0.77}
\definecolor{grey78}{rgb}{0.78,0.78,0.78}
\definecolor{grey79}{rgb}{0.79,0.79,0.79}
\definecolor{grey7}{rgb}{0.07,0.07,0.07}
\definecolor{grey80}{rgb}{0.80,0.80,0.80}
\definecolor{grey81}{rgb}{0.81,0.81,0.81}
\definecolor{grey82}{rgb}{0.82,0.82,0.82}
\definecolor{grey83}{rgb}{0.83,0.83,0.83}
\definecolor{grey84}{rgb}{0.84,0.84,0.84}
\definecolor{grey85}{rgb}{0.85,0.85,0.85}
\definecolor{grey86}{rgb}{0.86,0.86,0.86}
\definecolor{grey87}{rgb}{0.87,0.87,0.87}
\definecolor{grey88}{rgb}{0.88,0.88,0.88}
\definecolor{grey89}{rgb}{0.89,0.89,0.89}
\definecolor{grey8}{rgb}{0.08,0.08,0.08}
\definecolor{grey90}{rgb}{0.90,0.90,0.90}
\definecolor{grey91}{rgb}{0.91,0.91,0.91}
\definecolor{grey92}{rgb}{0.92,0.92,0.92}
\definecolor{grey93}{rgb}{0.93,0.93,0.93}
\definecolor{grey94}{rgb}{0.94,0.94,0.94}
\definecolor{grey95}{rgb}{0.95,0.95,0.95}
\definecolor{grey96}{rgb}{0.96,0.96,0.96}
\definecolor{grey97}{rgb}{0.97,0.97,0.97}
\definecolor{grey98}{rgb}{0.98,0.98,0.98}
\definecolor{grey99}{rgb}{0.99,0.99,0.99}
\definecolor{grey9}{rgb}{0.09,0.09,0.09}
\definecolor{grey}{rgb}{0.75,0.75,0.75}
\definecolor{honeydew1}{rgb}{0.94,1.00,0.94}
\definecolor{honeydew2}{rgb}{0.88,0.93,0.88}
\definecolor{honeydew3}{rgb}{0.76,0.80,0.76}
\definecolor{honeydew4}{rgb}{0.51,0.55,0.51}
\definecolor{honeydew}{rgb}{0.94,1.00,0.94}
\definecolor{hotpink}{rgb}{1.00,0.41,0.71}
\definecolor{indianred}{rgb}{0.80,0.36,0.36}
\definecolor{ivory1}{rgb}{1.00,1.00,0.94}
\definecolor{ivory2}{rgb}{0.93,0.93,0.88}
\definecolor{ivory3}{rgb}{0.80,0.80,0.76}
\definecolor{ivory4}{rgb}{0.55,0.55,0.51}
\definecolor{ivory}{rgb}{1.00,1.00,0.94}
\definecolor{khaki1}{rgb}{1.00,0.96,0.56}
\definecolor{khaki2}{rgb}{0.93,0.90,0.52}
\definecolor{khaki3}{rgb}{0.80,0.78,0.45}
\definecolor{khaki4}{rgb}{0.55,0.53,0.31}
\definecolor{khaki}{rgb}{0.94,0.90,0.55}
\definecolor{lavenderblush}{rgb}{1.00,0.94,0.96}
\definecolor{lavender}{rgb}{0.90,0.90,0.98}
\definecolor{lawngreen}{rgb}{0.49,0.99,0.00}
\definecolor{lemonchiffon}{rgb}{1.00,0.98,0.80}
\definecolor{lightblue}{rgb}{0.68,0.85,0.90}
\definecolor{lightcoral}{rgb}{0.94,0.50,0.50}
\definecolor{lightcyan}{rgb}{0.88,1.00,1.00}
\definecolor{lightgoldenrod}{rgb}{0.93,0.87,0.51}
\definecolor{lightgoldenrod}{rgb}{0.98,0.98,0.82}
\definecolor{lightgray}{rgb}{0.83,0.83,0.83}
\definecolor{lightgreen}{rgb}{0.56,0.93,0.56}
\definecolor{lightgrey}{rgb}{0.83,0.83,0.83}
\definecolor{lightpink}{rgb}{1.00,0.71,0.76}
\definecolor{lightsalmon}{rgb}{1.00,0.63,0.48}
\definecolor{lightsea}{rgb}{0.13,0.70,0.67}
\definecolor{lightsky}{rgb}{0.53,0.81,0.98}
\definecolor{lightslate}{rgb}{0.47,0.53,0.60}
\definecolor{lightslate}{rgb}{0.47,0.53,0.60}
\definecolor{lightslate}{rgb}{0.52,0.44,1.00}
\definecolor{lightsteel}{rgb}{0.69,0.77,0.87}
\definecolor{lightyellow}{rgb}{1.00,1.00,0.88}
\definecolor{limegreen}{rgb}{0.20,0.80,0.20}
\definecolor{linen}{rgb}{0.98,0.94,0.90}
\definecolor{magenta1}{rgb}{1.00,0.00,1.00}
\definecolor{magenta2}{rgb}{0.93,0.00,0.93}
\definecolor{magenta3}{rgb}{0.80,0.00,0.80}
\definecolor{magenta4}{rgb}{0.55,0.00,0.55}
\definecolor{magenta}{rgb}{1.00,0.00,1.00}
\definecolor{maroon1}{rgb}{1.00,0.20,0.70}
\definecolor{maroon2}{rgb}{0.93,0.19,0.65}
\definecolor{maroon3}{rgb}{0.80,0.16,0.56}
\definecolor{maroon4}{rgb}{0.55,0.11,0.38}
\definecolor{maroon}{rgb}{0.69,0.19,0.38}
\definecolor{mediumaquamarine}{rgb}{0.40,0.80,0.67}
\definecolor{mediumblue}{rgb}{0.00,0.00,0.80}
\definecolor{mediumorchid}{rgb}{0.73,0.33,0.83}
\definecolor{mediumpurple}{rgb}{0.58,0.44,0.86}
\definecolor{mediumsea}{rgb}{0.24,0.70,0.44}
\definecolor{mediumslate}{rgb}{0.48,0.41,0.93}
\definecolor{mediumspring}{rgb}{0.00,0.98,0.60}
\definecolor{mediumturquoise}{rgb}{0.28,0.82,0.80}
\definecolor{mediumviolet}{rgb}{0.78,0.08,0.52}
\definecolor{midnightblue}{rgb}{0.10,0.10,0.44}
\definecolor{mintcream}{rgb}{0.96,1.00,0.98}
\definecolor{mistyrose}{rgb}{1.00,0.89,0.88}
\definecolor{moccasin}{rgb}{1.00,0.89,0.71}
\definecolor{navajowhite}{rgb}{1.00,0.87,0.68}
\definecolor{navyblue}{rgb}{0.00,0.00,0.50}
\definecolor{navy}{rgb}{0.00,0.00,0.50}
\definecolor{oldlace}{rgb}{0.99,0.96,0.90}
\definecolor{olivedrab}{rgb}{0.42,0.56,0.14}
\definecolor{orange1}{rgb}{1.00,0.65,0.00}
\definecolor{orange2}{rgb}{0.93,0.60,0.00}
\definecolor{orange3}{rgb}{0.80,0.52,0.00}
\definecolor{orange4}{rgb}{0.55,0.35,0.00}
\definecolor{orangered}{rgb}{1.00,0.27,0.00}
\definecolor{orange}{rgb}{1.00,0.65,0.00}
\definecolor{orchid1}{rgb}{1.00,0.51,0.98}
\definecolor{orchid2}{rgb}{0.93,0.48,0.91}
\definecolor{orchid3}{rgb}{0.80,0.41,0.79}
\definecolor{orchid4}{rgb}{0.55,0.28,0.54}
\definecolor{orchid}{rgb}{0.85,0.44,0.84}
\definecolor{palegoldenrod}{rgb}{0.93,0.91,0.67}
\definecolor{palegreen}{rgb}{0.60,0.98,0.60}
\definecolor{paleturquoise}{rgb}{0.69,0.93,0.93}
\definecolor{paleviolet}{rgb}{0.86,0.44,0.58}
\definecolor{papayawhip}{rgb}{1.00,0.94,0.84}
\definecolor{peachpuff}{rgb}{1.00,0.85,0.73}
\definecolor{peru}{rgb}{0.80,0.52,0.25}
\definecolor{pink1}{rgb}{1.00,0.71,0.77}
\definecolor{pink2}{rgb}{0.93,0.66,0.72}
\definecolor{pink3}{rgb}{0.80,0.57,0.62}
\definecolor{pink4}{rgb}{0.55,0.39,0.42}
\definecolor{pink}{rgb}{1.00,0.75,0.80}
\definecolor{plum1}{rgb}{1.00,0.73,1.00}
\definecolor{plum2}{rgb}{0.93,0.68,0.93}
\definecolor{plum3}{rgb}{0.80,0.59,0.80}
\definecolor{plum4}{rgb}{0.55,0.40,0.55}
\definecolor{plum}{rgb}{0.87,0.63,0.87}
\definecolor{powderblue}{rgb}{0.69,0.88,0.90}
\definecolor{purple1}{rgb}{0.61,0.19,1.00}
\definecolor{purple2}{rgb}{0.57,0.17,0.93}
\definecolor{purple3}{rgb}{0.49,0.15,0.80}
\definecolor{purple4}{rgb}{0.33,0.10,0.55}
\definecolor{purple}{rgb}{0.63,0.13,0.94}
\definecolor{red1}{rgb}{1.00,0.00,0.00}
\definecolor{red2}{rgb}{0.93,0.00,0.00}
\definecolor{red3}{rgb}{0.80,0.00,0.00}
\definecolor{red4}{rgb}{0.55,0.00,0.00}
\definecolor{red}{rgb}{1.00,0.00,0.00}
\definecolor{rosybrown}{rgb}{0.74,0.56,0.56}
\definecolor{royalblue}{rgb}{0.25,0.41,0.88}
\definecolor{saddlebrown}{rgb}{0.55,0.27,0.07}
\definecolor{salmon1}{rgb}{1.00,0.55,0.41}
\definecolor{salmon2}{rgb}{0.93,0.51,0.38}
\definecolor{salmon3}{rgb}{0.80,0.44,0.33}
\definecolor{salmon4}{rgb}{0.55,0.30,0.22}
\definecolor{salmon}{rgb}{0.98,0.50,0.45}
\definecolor{sandybrown}{rgb}{0.96,0.64,0.38}
\definecolor{seagreen}{rgb}{0.18,0.55,0.34}
\definecolor{seashell1}{rgb}{1.00,0.96,0.93}
\definecolor{seashell2}{rgb}{0.93,0.90,0.87}
\definecolor{seashell3}{rgb}{0.80,0.77,0.75}
\definecolor{seashell4}{rgb}{0.55,0.53,0.51}
\definecolor{seashell}{rgb}{1.00,0.96,0.93}
\definecolor{sienna1}{rgb}{1.00,0.51,0.28}
\definecolor{sienna2}{rgb}{0.93,0.47,0.26}
\definecolor{sienna3}{rgb}{0.80,0.41,0.22}
\definecolor{sienna4}{rgb}{0.55,0.28,0.15}
\definecolor{sienna}{rgb}{0.63,0.32,0.18}
\definecolor{skyblue}{rgb}{0.53,0.81,0.92}
\definecolor{slateblue}{rgb}{0.42,0.35,0.80}
\definecolor{slategray}{rgb}{0.44,0.50,0.56}
\definecolor{slategrey}{rgb}{0.44,0.50,0.56}
\definecolor{snow1}{rgb}{1.00,0.98,0.98}
\definecolor{snow2}{rgb}{0.93,0.91,0.91}
\definecolor{snow3}{rgb}{0.80,0.79,0.79}
\definecolor{snow4}{rgb}{0.55,0.54,0.54}
\definecolor{snow}{rgb}{1.00,0.98,0.98}
\definecolor{springgreen}{rgb}{0.00,1.00,0.50}
\definecolor{steelblue}{rgb}{0.27,0.51,0.71}
\definecolor{tan1}{rgb}{1.00,0.65,0.31}
\definecolor{tan2}{rgb}{0.93,0.60,0.29}
\definecolor{tan3}{rgb}{0.80,0.52,0.25}
\definecolor{tan4}{rgb}{0.55,0.35,0.17}
\definecolor{tan}{rgb}{0.82,0.71,0.55}
\definecolor{thistle1}{rgb}{1.00,0.88,1.00}
\definecolor{thistle2}{rgb}{0.93,0.82,0.93}
\definecolor{thistle3}{rgb}{0.80,0.71,0.80}
\definecolor{thistle4}{rgb}{0.55,0.48,0.55}
\definecolor{thistle}{rgb}{0.85,0.75,0.85}
\definecolor{tomato1}{rgb}{1.00,0.39,0.28}
\definecolor{tomato2}{rgb}{0.93,0.36,0.26}
\definecolor{tomato3}{rgb}{0.80,0.31,0.22}
\definecolor{tomato4}{rgb}{0.55,0.21,0.15}
\definecolor{tomato}{rgb}{1.00,0.39,0.28}
\definecolor{turquoise1}{rgb}{0.00,0.96,1.00}
\definecolor{turquoise2}{rgb}{0.00,0.90,0.93}
\definecolor{turquoise3}{rgb}{0.00,0.77,0.80}
\definecolor{turquoise4}{rgb}{0.00,0.53,0.55}
\definecolor{turquoise}{rgb}{0.25,0.88,0.82}
\definecolor{violetred}{rgb}{0.82,0.13,0.56}
\definecolor{violet}{rgb}{0.93,0.51,0.93}
\definecolor{wheat1}{rgb}{1.00,0.91,0.73}
\definecolor{wheat2}{rgb}{0.93,0.85,0.68}
\definecolor{wheat3}{rgb}{0.80,0.73,0.59}
\definecolor{wheat4}{rgb}{0.55,0.49,0.40}
\definecolor{wheat}{rgb}{0.96,0.87,0.70}
\definecolor{whitesmoke}{rgb}{0.96,0.96,0.96}
\definecolor{white}{rgb}{1.00,1.00,1.00}
\definecolor{yellow1}{rgb}{1.00,1.00,0.00}
\definecolor{yellow2}{rgb}{0.93,0.93,0.00}
\definecolor{yellow3}{rgb}{0.80,0.80,0.00}
\definecolor{yellow4}{rgb}{0.55,0.55,0.00}
\definecolor{yellowgreen}{rgb}{0.60,0.80,0.20}
\definecolor{yellow}{rgb}{1.00,1.00,0.00}

\usepackage{amssymb}
\usepackage{amsmath}
\usepackage{amsthm}
\usepackage{bm}
\usepackage{stmaryrd}
\usepackage{accents}
\usepackage[mathscr]{eucal}
\usepackage[utf8]{inputenc}
\usepackage[english]{babel}

\newtheorem{definition}{Definition}
\newtheorem{remark}{Remark}

\newcommand{\notshow}[1]{}

\newcommand{\mat}[1]{\ensuremath{\mathbf{#1}}}                       
\newcommand{\ten}[1]{\ensuremath{\mathbf{\mathcal{#1}}}}           

\makeatletter
    \newcommand{\hadamard}{\mathbin{\mathpalette\make@circled\ast}}
    \newcommand{\make@circled}[2]{%
      \ooalign{$\m@th#1\smallbigcirc{#1}$\cr\hidewidth$\m@th#1#2$\hidewidth\cr}%
    }
    \newcommand{\smallbigcirc}[1]{%
      \vcenter{\hbox{\scalebox{0.77778}{$\m@th#1\bigcirc$}}}%
    }

    \newcommand{\circprod}[2]{\ensuremath{\,\underset{#1}{\overset{#2}{\circ}}}}

\makeatother

\newcommand{\rank}{\ensuremath{\,\textrm{rank}}}

\newcommand{\R}{\ensuremath{\mathbb{R}}}

\makeatletter
    \newcommand{\subalign}[1]{%
      \vcenter{%
        \Let@ \restore@math@cr \default@tag
        \baselineskip\fontdimen10 \scriptfont\tw@
        \advance\baselineskip\fontdimen12 \scriptfont\tw@
        \lineskip\thr@@\fontdimen8 \scriptfont\thr@@
        \lineskiplimit\lineskip
        \ialign{\hfil$\m@th\scriptstyle##$&$\m@th\scriptstyle{}##$\hfil\crcr
          #1\crcr
        }%
      }%
    }

\makeatother

\usepackage{graphicx}
\usepackage{caption}
\usepackage{subcaption}
\usepackage{rotating}

\graphicspath{%
    {02-figures/}%
}

\usepackage{tikz}
\usepackage{xifthen}
\usetikzlibrary{
    matrix,
    calc,
    positioning,
    shapes,shapes.geometric,
    arrows,arrows.meta,
    shadows,
    backgrounds,
    decorations.pathreplacing,calligraphy,
    decorations.markings,
    chains,
    mindmap,
    trees
}

\pgfdeclarelayer{background}
\pgfdeclarelayer{foreground}
\pgfsetlayers{background,main,foreground}

\newdimen\XCoord
\newdimen\YCoord
\newdimen\dummyCoord
\newlength\nodelength
\newlength\nodeheight
\newlength\nodeXshift
\newlength\figureHeight

\makeatletter
    \newcommand\getwidthofnode[2]{
        \pgfextractx{#1}{\pgfpointanchor{#2}{east}}
        \pgfextractx{\pgf@xa}{\pgfpointanchor{#2}{west}}
        \addtolength{#1}{-\pgf@xa}
    }

    \newcommand\getheightofnode[2]{
        \pgfextracty{#1}{\pgfpointanchor{#2}{south}}
        \pgfextracty{\pgf@xa}{\pgfpointanchor{#2}{north}}
        \addtolength{#1}{-\pgf@xa}
    }
\makeatother

\tikzset{
    invisible/.style = {
        opacity=0
    },
    visible on/.style = {alt={#1{}{invisible}}},
    alt/.code args = {<#1>#2#3}{ \alt<#1>{\pgfkeysalso{#2}}{\pgfkeysalso{#3}}},
    arrow/.style = {-{Stealth[scale=1.5]}, rounded corners, very thick},
    circle dotted/.style={
        dash pattern=on 0.5pt off 10pt,
        line cap=round,
        line width=2pt
    },
    mydash/.style={
        line width=3pt,
        dashed,
        color=SeaGreen2,
        dash pattern=on 15pt off 15pt,
    },
    every node/.style = {
        anchor=south west
    },
    my_matrix/.style = {
        matrix,
        fill=white,
        anchor=west,
        left delimiter={[},
        right delimiter={]},
        row sep=-1pt,
        column sep=4mm,
        outer sep=0pt
    },
    my_matrix2/.style = {
        matrix,
        fill=white,
        anchor=west,
        row sep=-1pt,
        column sep=4mm,
        outer sep=0pt
    },
    no sep/.style={
        inner sep=0pt,
        outer sep=0pt
    },
    double line/.style={
        line width=5pt,
        double distance=10pt,
    },
    cross/.style={
        path picture={
            \draw[black]
                (path picture bounding box.south east) -- (path picture bounding box.north west)
                (path picture bounding box.south west) -- (path picture bounding box.north east);
        }
     },
    kronecker/.style={
        draw,
        line width=3pt,
        fill=white,
        circle,
        cross,
        minimum size=1cm,
        anchor=center
    },
    unfolding/.style={
        line width=3pt,
        draw,
        fill=white,
        cross,
        minimum size=0.8cm,
        anchor=center
    },
    round corners/.style={
        line width=3pt,
        rounded corners=5mm
    },
}

\pgfkeys{
    /tensorParts/.is family, /tensorParts,
    default/.style={
        width =3,
        height=3,
        depth=3,
        start at={0,0},
        xshift=0,
        yshift=0,
        zshift=0,
        color frontal = red,
        color lateral = blue,
        color horizontal = yellow,
        color = red,
        inner color = white,
        anchor = south west,
        name = dummy,
        line width = 0.01mm
    },
    width/.estore in = \mywidth,
    height/.estore in = \myheight,
    depth/.estore in = \mydepth,
    start at/.estore in = \mystartat,
    xshift/.estore in = \myxshift,
    yshift/.estore in = \myyshift,
    zshift/.estore in = \myzshift,
    color frontal/.estore in = \mycolorfrontal,         
    color lateral/.estore in = \mycolorlateral,         
    color horizontal/.estore in = \mycolorhorizontal,   
    color/.estore in = \mycolor,                        
    inner color/.estore in = \myinnercolor,
    anchor/.estore in = \myanchor,                      
    name/.estore in = \myname,
    line width/.estore in = \mylinewidth,
    /Tensor/.is family, /Tensor,
    defaultTensor/.style={
        width=3,
        height=3,
        depth=3,
        start at={0,0},
        xshift=0,
        yshift=0,
        zshift=0,
        colorf=colorFrontal,
        colorl=colorLateral,
        colorh=colorHorizontal,
        color=white,
        inner color=white,
        fill opacity=1,
        anchor=south west,
        name=dummy,
        line width=3.5pt,
        on slide=1-
    },
    width/.estore in = \width,
    height/.estore in = \height,
    depth/.estore in = \depth,
    start at/.estore in = \startat,
    xshift/.estore in = \xshift,
    yshift/.estore in = \yshift,
    zshift/.estore in = \zshift,
    colorf/.estore in = \tcolorf,             
    colorl/.estore in = \tcolorl,             
    colorh/.estore in = \tcolorh,             
    color/.estore in = \tcolor,               
    inner color/.estore in = \innercolor,
    anchor/.estore in = \anchor,             
    fill opacity/.estore in = \fillopacity,
    name/.estore in = \name,
    line width/.estore in = \linewidth,
    on slide/.estore in = \onslide,
    /SimplePartsTNs/.is family, /SimplePartsTNs,
    defaultSimplePartsTNs/.style={
        inner color=white,
        tnode color=tnode,
        tnode size=3cm,
        tnode anchor=center,
        mnode color=mnode,
        mnode size=2cm,
        mnode anchor=center,
        leaf size=5cm,
        leaf width=3pt,
        bar size=30pt,
        bar width=2pt,
        raise label=15pt
    },
    inner color/.estore in = \innercolor,
    tnode color/.estore in = \tnodecolor,
    tnode size/.estore in = \tnodesize,
    tnode anchor/.estore in = \tnodeanchor,
    mnode color/.estore in = \mnodecolor,
    mnode size/.estore in = \mnodesize,
    mnode anchor/.estore in = \mnodeanchor,
    leaf size/.estore in = \leafsize,
    leaf width/.estore in = \leafwidth,
    bar size/.estore in = \barsize,
    bar width/.estore in = \barwidth,
    raise label/.estore in = \raiselabel,
    /TNs/.is family, /TNs,
    defaultTNs/.style={
        name=dummy,
        start at={0,0},
        xshift=0,
        yshift=0,
        label={},
        type=e,
        layer=foreground,
    },
    name/.estore in = \name,
    start at/.estore in = \startat,
    xshift/.estore in = \xshift,
    yshift/.estore in = \yshift,
    label/.estore in = \label,
    type/.estore in = \type,
    layer/.estore in = \layer,
}

\newcommand\tnLeaf[4][]{
    \pgfkeys{/TNs, defaultTNs, #1}
        \begin{pgfonlayer}{background}
        \node[leaf-\type, #2] (\name) at ($(\startat)+(\xshift,\yshift)$){};
        \node[bar-\type, #3] () at (\name){};
        \node[dim-\type, #4] () at (\name){\label};
        \end{pgfonlayer}
}

\newcommand\tnLeafCC[5][]{
    \ifthenelse{\isempty{#5}}%
    {\tnLeaf[type=c, #1]{#2}{#3}{#4}}%
    {\tnLeaf[type=c, #1]{minimum width=#5, #2}{#3}{#4}}%
}

\newcommand\tnLeafEE[5][]{
    \ifthenelse{\isempty{#5}}%
    {\tnLeaf[type=e, #1]{#2}{#3}{#4}}%
    {\tnLeaf[type=e, #1]{minimum width=#5, #2}{#3}{#4}}%
}

\newcommand\tnLeafWW[5][]{
    \ifthenelse{\isempty{#5}}%
    {\tnLeaf[type=w, #1]{#2}{#3}{#4}}%
    {\tnLeaf[type=w, #1]{minimum width=#5, #2}{#3}{#4}}%
}

\newcommand\tnLeafNN[5][]{
    \ifthenelse{\isempty{#5}}%
    {\tnLeaf[type=n, #1]{#2}{#3}{#4}}%
    {\tnLeaf[type=n, #1]{minimum height=#5, #2}{#3}{#4}}%
}

\newcommand\tnLeafNE[5][]{
    \ifthenelse{\isempty{#5}}%
    {\tnLeaf[type=ne, #1]{#2}{#3}{#4}}%
    {\tnLeaf[type=ne, #1]{minimum height=#5, #2}{#3}{#4}}%
}

\newcommand\tnLeafNW[5][]{
    \ifthenelse{\isempty{#5}}%
    {\tnLeaf[type=nw, #1]{#2}{#3}{#4}}%
    {\tnLeaf[type=nw, #1]{minimum height=#5, #2}{#3}{#4}}%
}

\newcommand\tnLeafSS[5][]{
    \ifthenelse{\isempty{#5}}%
    {\tnLeaf[type=s, #1]{#2}{#3}{#4}}%
    {\tnLeaf[type=s, #1]{minimum height=#5, #2}{#3}{#4}}%
}

\newcommand\tnLeafSE[5][]{
    \ifthenelse{\isempty{#5}}%
    {\tnLeaf[type=se, #1]{#2}{#3}{#4}}%
    {\tnLeaf[type=se, #1]{minimum width=#5, #2}{#3}{#4}}%
}

\newcommand\tnLeafSW[5][]{
    \ifthenelse{\isempty{#5}}%
    {\tnLeaf[type=sw, #1]{#2}{#3}{#4}}%
    {\tnLeaf[type=sw, #1]{minimum width=#5, #2}{#3}{#4}}%
}

\usepackage{tcolorbox}
\usepackage{tabularx}
\usepackage{colortbl}
\usepackage{multirow}
\usepackage{booktabs}
\usepackage[framemethod=TikZ]{mdframed}

\tcbuselibrary{skins}
\newcolumntype{Y}{>{\raggedleft\arraybackslash}X}

\tcbset{
    tab1/.style={
        fonttitle=\bfseries\large,
        fontupper=\normalsize\sffamily,
        colback=white,colframe=red!75!black,
        colbacktitle=Salmon!40!white,
        coltitle=black,
        center title,
        freelance,
        frame code={
            \foreach \n in {north east,north west,south east,south west}{
                \path [fill=red!75!black] (interior.\n) circle (3mm);
            };
        },
    },
    tab2/.style={
        enhanced,
        fonttitle=\bfseries,
        fontupper=\normalsize\sffamily,
        colback=white,
        colframe=blue!50!black,
        colbacktitle=Salmon!40!white,
        coltitle=black,
        center title,
    },
}

\newtcolorbox[blend into=tables]{mytable}[2][]{
    tab2,
    float=tbh,
    halign=center,
    title={#2},
    every float=\centering,
    #1
}

\mdfdefinestyle{mdf-figures}{
    linecolor=blue!50!black,
    backgroundcolor=white,
    roundcorner=8pt,
    middlelinewidth=1.5pt
}

\renewenvironment{figure*}[1][]{
    \begin{originalfigure}[#1]
        \begin{mdframed}[style=mdf-figures]
            \centering
            \footnotesize
        }{
        \end{mdframed}
    \end{originalfigure}
}
\renewenvironment{figure}[1][]{
    \begin{originalfigure}[#1]
        \begin{mdframed}[style=mdf-figures]
            \centering
            \footnotesize
        }{
        \end{mdframed}
    \end{originalfigure}
}

\renewenvironment{sidewaysfigure}[1][]{
    \begin{originalSidewaysFigure}[#1]
        \begin{mdframed}[style=mdf-figures]
            \centering
            \footnotesize
        }{
        \end{mdframed}
    \end{originalSidewaysFigure}
}

\usepackage{amsmath,amsfonts}
\usepackage{algorithmic}
\usepackage{array}
\usepackage{textcomp}
\usepackage{stfloats}
\usepackage{url}
\usepackage{verbatim}
\usepackage{graphicx}
\usepackage{cite}
\usepackage{hyperref}
\usepackage{booktabs}
\usepackage{subcaption}
\usepackage{caption}
\usepackage{tikz}
\usepackage[export]{adjustbox}
\usepackage[margin=1in]{geometry}
\newtheorem{thm}{Observation}

\begin{document}

\title{Understanding the Rank of Tensor Networks via an Intuitive Example-Driven Approach}
%
\author{Wuyang~Zhou,~\IEEEmembership{Graduate Student Member,~IEEE}, Giorgos~Iacovides,~\IEEEmembership{Graduate Student Member,~IEEE}, Kriton~Konstantinidis, Ilya~Kisil,
        and~Danilo~P.~Mandic,~\IEEEmembership{Fellow,~IEEE}

\thanks{All authors are with the Department of Electrical and Electronic Engineering, Imperial College London, London SW7 2AZ, U.K. (e-mails: \{wuyang.zhou19, giorgos.iacovides20, k.konstantinidis19, i.kisil15, d.mandic\}@imperial.ac.uk) }
}
%
\markboth{A preprint}%
{Zhou \MakeLowercase{\textit{et al.}}: Understanding the Rank of Tensor Networks via an Intuitive Example-Driven Approach}

\maketitle
\vspace{-14mm}
%
%
\IEEEPARstart{T}{ensor} Network (TN) decompositions have emerged as an indispensable tool in Big Data analytics owing to their ability to provide compact low-rank representations, thus alleviating the ``Curse of Dimensionality'' inherent in handling higher-order data. At the heart of their success lies the concept of TN ranks, which governs the efficiency and expressivity of TN decompositions. However, unlike matrix ranks, TN ranks often lack a universal meaning and an intuitive interpretation, with their properties varying significantly across different TN structures. Consequently, TN ranks are frequently treated as empirically tuned hyperparameters, rather than as key design parameters inferred from domain knowledge. The aim of this Lecture Note is therefore to demystify the foundational yet frequently misunderstood concept of TN ranks through real-life examples and intuitive visualizations. We begin by illustrating how domain knowledge can guide the selection of TN ranks in widely-used models such as the Canonical Polyadic (CP) and Tucker decompositions. For more complex TN structures, we employ a self-explanatory graphical approach that generalizes to tensors of arbitrary order. Such a perspective naturally reveals the relationship between TN ranks and the corresponding ranks of tensor unfoldings (matrices), thereby circumventing cumbersome multi-index tensor algebra while facilitating domain-informed TN design. It is our hope that this Lecture Note will equip readers with a clear and unified understanding of the concept of TN rank, along with the necessary physical insight and intuition to support the selection, explainability, and deployment of tensor methods in both practical applications and educational contexts.

\vspace{-2mm}
\section*{Scope and Relevance}
\vspace{-1mm}

The rapid growth in both the volume and diversity of generated data has highlighted the need for more versatile and computationally efficient data analysis tools, beyond the standard ``flat-view'' matrix approaches rooted in linear algebra. Tensors, or multi-way arrays, provide such an algebraic framework which is inherently suited to data of higher-order, large volume, and diversity. However, higher-order tensors in their original format have a storage complexity which scales exponentially with the number of tensor dimensions (modes), a phenomenon referred to as the ``Curse of Dimensionality" \cite{Cichocki2014TensorDF, TT_decomposition}. This issue can be greatly mitigated through TN decompositions, which represent the original tensor as a set of smaller-sized factors connected via tensor contractions \cite{Cichocki2014TensorDF}. Indeed, TN methods have achieved remarkable success in dimensionality reduction across diverse domains, such as machine learning, signal processing, computer vision, and quantum physics \cite{Cichocki2014TensorDF, tensorization, TT_decomposition, orus2014practical}. Widely used TN structures include the Canonical Polyadic (CP) decomposition, Tucker decomposition, and Tensor Train (TT) decomposition \cite{kolda2009tensor, TT_decomposition}.

A key property underpinning any Tensor Network (TN) structure is the notion of tensor rank (TN rank). In the matrix case, rank is well-defined and is directly connected to fundamental linear algebraic properties, such as the dimension of row or column spaces. In contrast, TN ranks lack a consistent or ``universal'' interpretation across different TN structures, forcing practitioners to treat them as empirically chosen hyperparameters. Moreover, determining the optimal rank for higher-order tensors is NP-hard and thus often computationally intractable in practice \cite{tnale_cite}. The derivation of rank-related properties for specific TN formats also typically involves advanced concepts from multilinear algebra \cite{Zheng_Huang_Zhao_Zhao_Jiang_2021, domanov2021computation, HOSVD}, which can pose a barrier for practitioners in Big Data analytics who lack specialized training in tensor decompositions. As a result, the selection and design of suitable TN structures for applications such as tensor compression often becomes a trial-and-error process, rather than being guided by a principled framework informed by domain knowledge and multilinear rank theory.

This Lecture Note aims to demystify the concept of TN ranks for a generally knowledgeable reader by revisiting TN ranks via an intuitive example-driven approach. This is achieved by:
\begin{enumerate}
\item Leveraging the rank properties of the Canonical Polyadic (CP) decomposition to promote the explainability and uniqueness of the CP factors, as exemplified through strict frequency-channel separation in the fundamental paradigm of time-frequency representation (TFR). 
\item Demonstrating, through the lens of domain knowledge, how varying the ranks in the CP and Tucker decompositions influences the expressivity of higher-order TN representations, as illustrated via intuitive examples involving RGB color images.
\item Employing a diagrammatic procedure to link the rank properties of tensor unfoldings (matrices) to TN ranks, without the need to resort to cumbersome multilinear algebra. This facilitates the design of custom TN structures, applies to tensors of arbitrary order, and can be performed directly on the TN graph, as demonstrated through an example on RGB color video compression.
\end{enumerate}
It is our hope that the so introduced transparency and ease of interpretation will not only help promote informed selection of TN ranks (rather than relying on heuristic tuning), but will also demystify TN ranks for educational purposes, while empowering practitioners with greater intuition and design freedom.

\section*{Prerequisites}
\begin{figure}
\centering
\includegraphics[width=1\columnwidth]{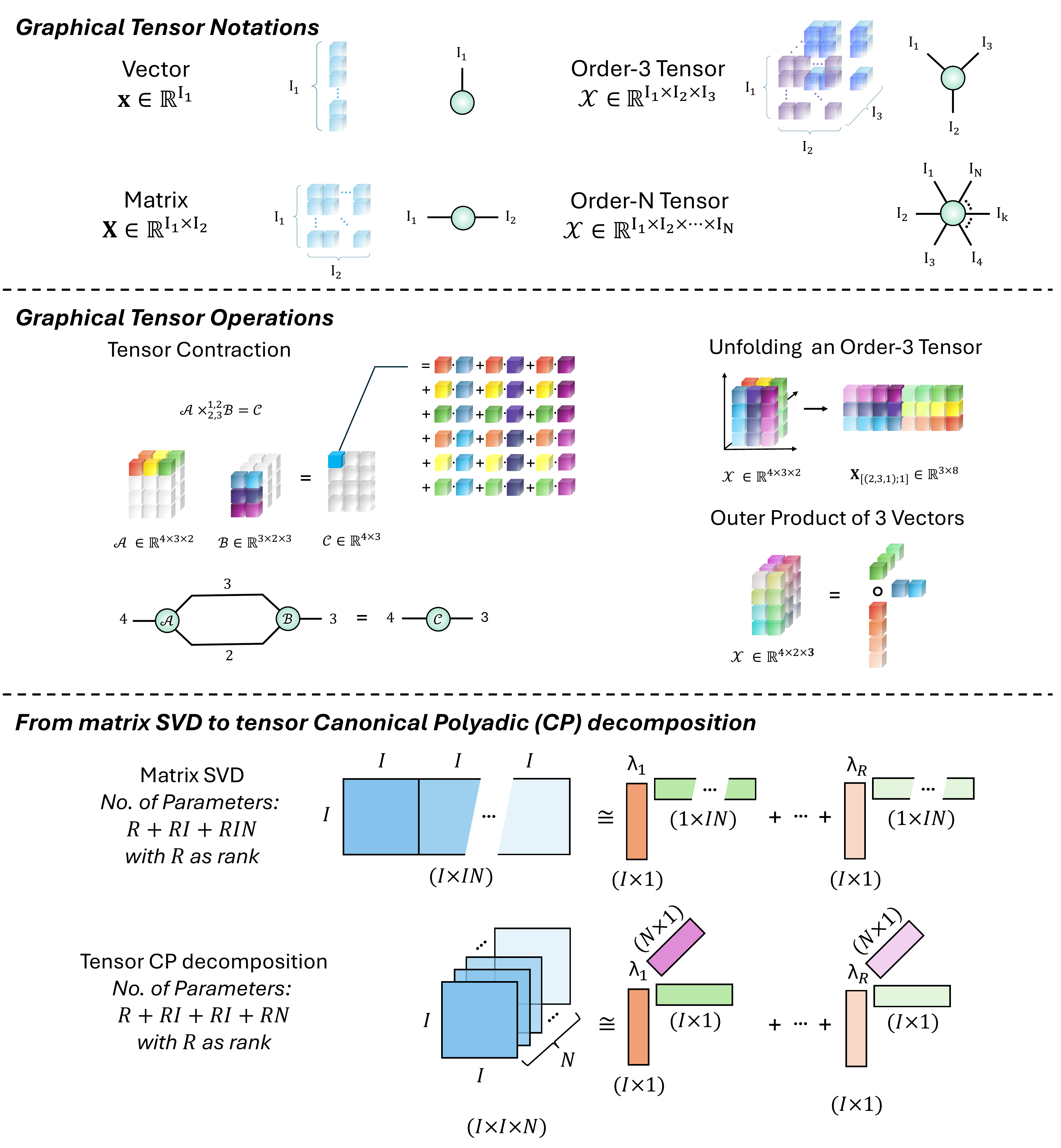}
\caption{Tensor notations, operations and multi-linear decompositions. \textbf{Top:} Graphical notations for a vector, a matrix, and higher-order tensors. \textbf{Middle:} Basic operations on tensor networks, including tensor contraction, unfolding (matricization), and outer product.  \textbf{Bottom:} Canonical Polyadic (CP) decomposition may be interpreted as an extension of the matrix singular value decomposition (SVD) to higher-order tensors, which approximates the original tensor by a set of rank-$1$ tensors of the same dimensionality. Observe that the tensor CP decomposition achieves superior compression compared to the matrix SVD for $J+K < JK$, which is especially prominent for large mode sizes $J$ and $K$ (see also Supplementary Material).}
\label{fig:Tensor_notations}
\end{figure}

This Lecture Note assumes a basic knowledge of Linear Algebra and Digital Signal Processing (DSP), such as Fourier transform and time-frequency representation.

\subsection*{Background}
Scalars, vectors, matrices and order-$N$ tensors are denoted respectively by lowercase letters, $x \in \mathbb{R}^1$, bold lowercase letters, $\mathbf{x} \in \mathbb{R}^{I_1}$, bold uppercase letters, $\mathbf{X} \in \mathbb{R}^{I_1 \times I_2}$, and calligraphic uppercase letters, $\mathcal{X} \in \mathbb{R}^{I_1 \times I_2 \times \ldots \times I_N}$, as shown in Figure \ref{fig:Tensor_notations} (Top). Moreover, we consider a matrix $\mathbf{X} \in \mathbb{R}^{I_1 \times  \prod_{n=2}^N I_n }$ as an order-$2$ tensor and  $\mathcal{X} \in \mathbb{R}^{I_1 \{\times I_n\}_{n=2}^N }$ as an order-$N$ tensor. The $(i_1, i_2, \ldots, i_N)$th-element of a tensor $\mathcal{X}$ is denoted by $\mathcal{X}(i_1, i_2, \ldots, i_N)$. The Frobenius norm is denoted by $\| \cdot \|_F$, while the element-wise definition of an \textit{outer product}  of $N$ vectors is given by
\begin{equation}
    \ten{X}(i_1,i_2, \ldots, i_N) = \prod_{n=1}^{N} \mat{x}^{(n)}(i_n)
\end{equation}
\begin{definition}
    \label{def:rank-1}
    If a tensor, $\ten{X} \in \R^{I_1 \times \cdots \times I_N}$, can be expressed as an
    outer product of $N$ vectors $\mat{x}^{(n)} \in \R^{I_n} , \ \forall n$, that is
    \begin{equation}
        \ten{X} = \mat{x}^{(1)} \circ \mat{x}^{(2)} \circ \cdots \circ \mat{x}^{(N)}
    \end{equation}
    then such multi-dimensional array (tensor) is said to be of rank-1.
\end{definition}

\begin{definition}
A permutation of the mode indices in a tensor $\mathcal{X}$ according to a permutation vector $\mathbf{p} = ( p_1, p_2, \ldots, p_N )$ is denoted as $\overrightarrow{\mathcal{X}^{\mathbf{p}}} \in \mathbb{R}^{I_{p_1} \times I_{p_2} \times \cdots \times I_{p_N}}$. 
\end{definition}
\begin{definition}
Unfolding operations convert tensors into matrices or vectors and are defined as 
\begin{equation}
\mathbf{X}_{[\mathbf{p};d]}(\overline{i_{p_1} \cdots i_{p_d}}, \overline{i_{p_{d+1}} \cdots i_{p_N}})\in \mathbb{R}^{\prod_{i = 1}^d I_{p_i} \times \prod_{j = d+1}^N I_{p_j}} = \overrightarrow{\mathcal{X}^{\mathbf{p}}}(i_{p_1}, i_{p_2}, \ldots, i_{p_N}) 
\end{equation}
where $1 \leq d \leq N-1$ (see also Figure \ref{fig:Tensor_notations} (Middle)). The specific unfolding case with $d=1$ is referred to as the mode-$k$ unfolding $(k=p_1)$ \cite{domanov2021computation}. The inverse operation of unfolding is termed folding or tensorization, and is denoted by $\text{Fold}_{[\mathbf{p};d]}$\cite{Zheng_Huang_Zhao_Zhao_Jiang_2021, tensorization}. 
\end{definition}

\begin{definition}
    Tensor contractions between two tensors are only allowed between modes of common sizes, as shown in Figure \ref{fig:Tensor_notations} (Middle). For example, to contract the order-$N$ tensor, $\mathcal{X}\in \mathbb{R}^{I_1 \times I_2 \times \cdots \times I_N}$, with the order-$M$ tensor, $\mathcal{Y}\in \mathbb{R}^{J_1 \times J_2 \times \cdots \times J_M}$, we first let the two permutation vectors, $\mathbf{p^x}$ and $\mathbf{p^y}$, have their first $k \left( 1 \leq k \leq \operatorname{min}\left(N,M\right)\right)$ elements correspond to the $k$ common-sized modes in $\mathcal{X}$ and $\mathcal{Y}$, while the other entries follow their original ascending order. A tensor contraction is then given by
\begin{equation}
\begin{split}
    \mathcal{Q}&= \mathcal{X} \times^{{\mathbf{p^y}(1), \mathbf{p^y}(2), \ldots, \mathbf{p^y}(k)}}_{\mathbf{p^x}(1), \mathbf{p^x}(2), \ldots, \mathbf{p^x}(k)} \mathcal{Y} = \text{Fold}_{[(1,2,\ldots, M+N-2k);M-k]} (\mathbf{X}^{\mathsf{T}}_{[\mathbf{p^x};k]}  \mathbf{Y}_{[\mathbf{p^y};k]}) \\
    &\in \mathbb{R}^{I_{\mathbf{p^x}(k+1)} \times I_{\mathbf{p^x}(k+2)} \times \cdots \times I_{\mathbf{p^x}(N)} \times J_{\mathbf{p^y}(k+1)} \times J_{\mathbf{p^y}(k+1)} \times \cdots \times J_{\mathbf{p^y}(M)}}
    \label{equ:tensor_contraction}
\end{split}
\end{equation}
The element-wise definition of the so obtained tensor $\mathcal{Q}$ is given by 
\begin{equation}
\begin{split}
    &\mathcal{Q}(i_{p_{k+1}}, i_{p_{k+2}}, \ldots, i_{p_N},j_{p_{k+1}}, j_{p_{k+2}}, \ldots, j_{p_M}) \\
    &\quad = \overrightarrow{\mathcal{X}^{\mathbf{p^x}}}(\underbrace{:,:,\ldots,,:}_{k \text{ times}},i_{p_{k+1}}, i_{p_{k+2}}, \ldots, i_{p_N} ) \overrightarrow{\mathcal{Y}^{\mathbf{p^y}}}(\underbrace{:,:,\ldots,,:}_{k \text{ times}},j_{p_{k+1}}, j_{p_{k+2}}, \ldots, j_{p_M} ) \\
\end{split}
\end{equation}
\end{definition}
\subsection*{Canonical Polyadic (CP) Decompositon} Following from Definition \ref{def:rank-1} and by analogy with linear algebra, a tensor, $\ten{X} \in \R^{I_1 \times
\cdots \times I_N}$, is said to be of rank-$R$ if it can be represented as a linear combination of $R$ rank-$1$ tensors in the form
\begin{equation}
    \ten{X} = \sum_{r=1}^R \ten{A}_r = \sum_{r=1}^R \mat{a}_r^{(1)} \circ \mat{a}_r^{(2)} \circ \cdots \circ \mat{a}_r^{(N)} = \sum_{r=1}^R \circprod{n=1}{N} \mat{a}_r^{(n)}.
\end{equation}

The core idea behind the CP Decomposition, as illustrated in Figure \ref{fig:Tensor_notations} (Bottom), is to decompose an $N$-dimensional array into a sum of rank-$1$ tensor factors \cite{6544287}, each weighted by a scalar, $\sigma_r$, to give
\begin{equation}
 \label{eq.CPD1}
 \ten{X} = \sum_{r = 1}^{R} \sigma_r \cdot \mathbf{a}_r^{(1)} \circ \mathbf{a}_r^{(2)} \circ \cdots \circ \mathbf{a}_r^{(N)}= \sum_{r=1}^{R} \ten{G}(r, \ldots, r) \mat{A}^{(1)}(i_1, r) \mat{A}^{(2)}(i_2, r) \cdots\mat{A}^{(N)}(i_N, r)
\end{equation}
where $\ten{G}$ is a \textit{super-diagonal} core tensor, and $\mat{A}^{(n)}$ are the factor matrices which comprise the corresponding vectors $\mat{a}_r^{(n)}$ for $ r = 1, \ldots, R$ and $n = 1, \ldots, N$. 

Since the individual vector terms in each rank-$1$ tensor factor do not interact with the vector terms in other tensor factors, the TN rank of the CP decomposition (CP rank) is $R$, that is, the number of rank-1 tensors that is summed up to yield $\mathcal{X}$. 

\par
\noindent\textbf{Uniqueness.} The uniqueness condition of the order-$3$ CP decomposition relates the CP rank $R$ and the ranks of the factor matrices in Equation (\ref{eq.CPD1}), and is given by the so called Kruskal uniqueness condition
\begin{gather}
   \label{eq.uniq1}
   \rank (\mat{A}^{(1)}) + \rank (\mat{A}^{(2)}) + \rank(\mat{A}^{(3)}) \geq 2R +2
\end{gather}

For a tensor of order-$4$ and greater, the corresponding sufficient condition for the CP decomposition of an order-$N$ tensor to be unique is given by \cite{kolda2009tensor}
\begin{equation}
   \sum_{n=1}^N \rank (\mat{A}^{(n)}) \ge 2R + N-1
   \label{eq.uniq2}
\end{equation}

\begin{remark}
    There is a subtle distinction between the role of rank in TN decompositions and the rank in traditional matrix methods such as Principal Component Analysis (PCA). In PCA, the principal components are obtained sequentially, so the first component remains unchanged whether one or many components are extracted. In contrast, the factors of tensor decompositions are usually computed jointly. Consequently, changing the rank for a CP decomposition algorithm from $R$ to $R+1$ will typically change all of the previously computed CP factors for rank-$R$. Thus, in the absence of domain knowledge, the TN rank is usually either given by: i) a best guess; ii) dictated by computational constraints; or iii) determined by trial-and-error. This signifies the importance of understanding TN ranks in order to fully utilize TN methods. This is elaborated further in the Supplementary Material.
\end{remark}

\section*{Problem statement and solution}
We next illustrate how domain knowledge can be used to guide informed selection of TN ranks.

\begin{figure}[htbp]
    \centering
     \includegraphics[width=1\textwidth]{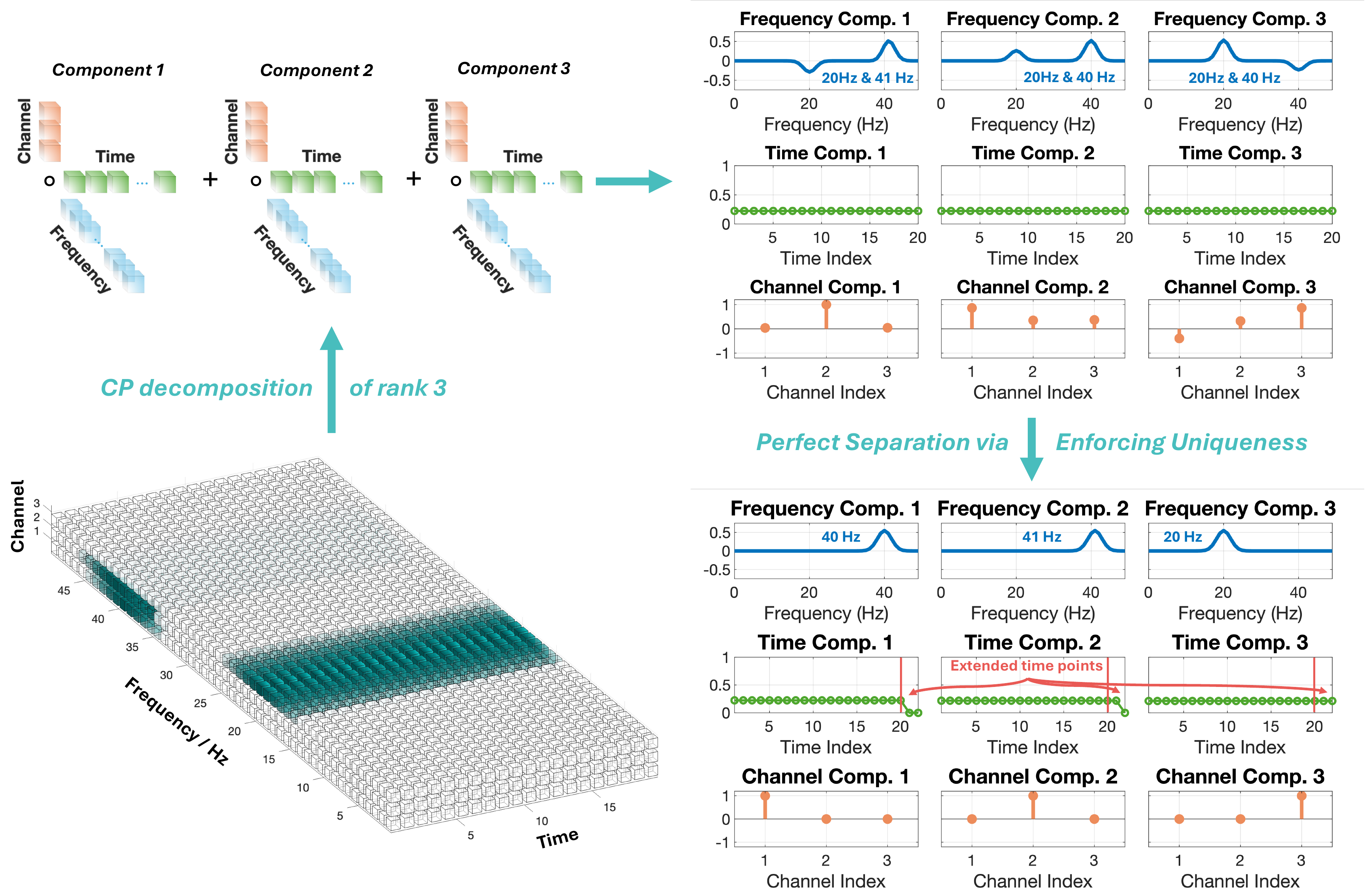};
      \caption{
          Explainability of CP decomposition, illuminated through an example of Time-Frequency Representation (TFR), whereby strict frequency-channel separation can only be achieved by enforcing uniqueness into CP decomposition. 
          \textbf{Bottom Left:} The order-$3$ \textit{frequency} $\times$ \textit{time} $\times$ \textit{channel} TFR tensor in the original format. Each data channel (signal) contains only one dominant sinusoidal frequency at $20$ Hz, $40$ Hz, or $41$ Hz.
          \textbf{Top Left:} CP decomposition of rank-$3$ decomposes the TFR tensor into $3$ rank-$1$ factors.
          \textbf{Top Right:} Applying the CP decomposition naïvely leads to overlapped frequency components in the CP factors (in blue), and mixed channel contributions in the channel components (in orange), as e.g., all three frequency factors contain $20$ Hz frequencies and two factors contain the $40$ Hz frequency. This  cross-channel interference leads to the consequent lack of explainability of the CP components.
          \textbf{Bottom Right:} By augmenting the time dimension of the TFR tensor to meet the Kruskal uniqueness condition in Equation (\ref{eq.proof5}), we can guarantee a perfect frequency separation of the original TFR tensor in the individual factors. This is reflected in the frequency components, each of which contains only a single dominant frequency, while the channel components each exhibit only a single non-zero channel index.}
     \label{fig:tf_cp}
  \end{figure}
\vspace{-3mm}
\subsection*{Elucidating TN ranks through Time Frequency Representation of Signals}
\label{sec.fakeTFRproblem}

The CP decomposition is a well-established method for efficient low-rank representations of data; however, such tensor-domain representation often violates the uniqueness conditions in Equation (\ref{eq.uniq1}) and (\ref{eq.uniq2}). To this end, we next employ a practical data augmentation procedure, which ensures the uniqueness of the CP decomposition. For enhanced intuition, this is achieved in the context of Time-Frequency Representation (TFR) \cite{miwakeichi2004decomposing}, where our objective is to construct an explainable model that decomposes a three-way TFR tensor, shown in Figure \ref{fig:tf_cp}, into a sum of rank-$1$ components, each corresponding to a single dominant frequency in every data channel.

\vspace{-3mm}
\subsection*{Example 1: TN ranks, Data Augmentation, and Model Explainability}
An order-$3$ TFR tensor was constructed by applying the Short-Time Fourier Transform (STFT) to each row of a data matrix representing the time-domain amplitudes of three sinusoidal signals at $20$ Hz, $40$ Hz, and $41$ Hz. Given the three frequency components, it is natural to factorize the resulting tensor via a CP decomposition of rank $3$ into its elementary factors, whereby the CP rank determines the number of such fundamental components. The CP decomposition of rank-$3$ using the alternating least squares (ALS) algorithm \cite{kolda2009tensor} yields the results shown in Figure \ref{fig:tf_cp} (Top Right).

\begin{remark}
Observe from Figure \ref{fig:tf_cp} (Top Right) that a naïve use of the CP decomposition leads to interferences in each CP frequency component (in blue) and mixed channel contributions in the CP channel components (in red), leading to a lack of explainability in the CP factors. This is exemplified by the presence of $20$ Hz frequency in all three frequency factors in Figure \ref{fig:tf_cp} (Top Right). Moreover, every run of the CP-ALS algorithm produces a different set of CP components due to the randomness in the ALS optimization process, further preventing the identifiability of naïve CP decomposition. Therefore, even if perfect separation may occur by chance, the uniqueness of such CP decomposition is not inherently guaranteed. This underscores the importance of using the CP rank properties to enforce identifiability.
\end{remark}

\begin{figure*}
   \centering
    \includegraphics[width=\textwidth]{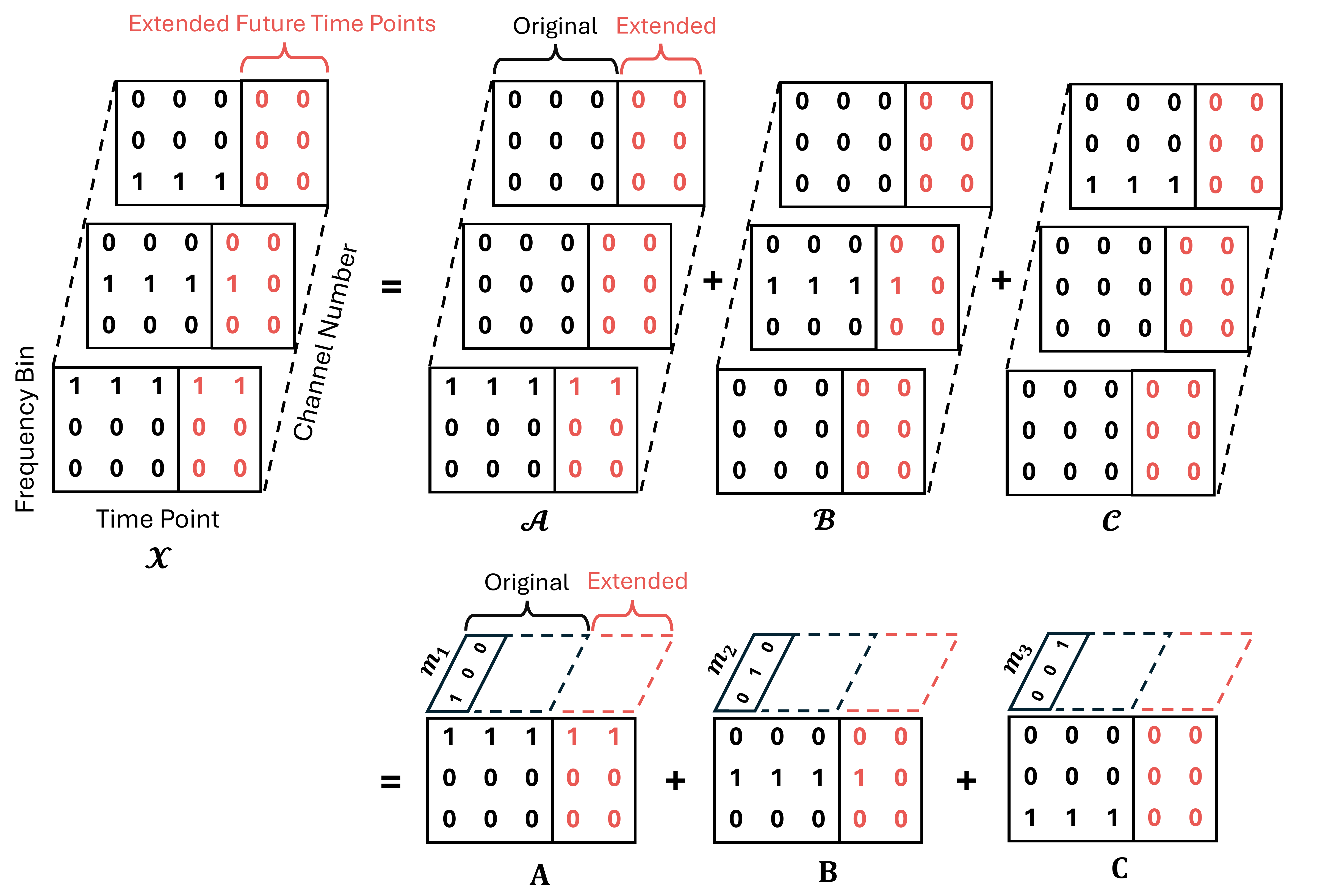}
   \caption{Data augmentation of the ideal TFR tensor from Equation (\ref{eq.proof1}), which makes all factor matrices have full rank and hence satisfy the Kruskal uniqueness condition in Equation (\ref{eq.proof5}). The resulting decomposition yields three rank-$1$ tensors (top panel)  with distinct frequency components. To ensure the uniqueness of CP decomposition, we extend (augment) the time mode, with the corresponding changes in the factors designated in red. The bottom panel shows the base matrices (\( \mat{A}, \mat{B}, \text{and } \mat{C}\)) of each individual channel and the channel mode vectors \( \mat{m}_1 \), \( \mat{m}_2 \), and \( \mat{m}_3 \).}
   \label{fig.originalFakeTensor}
\end{figure*}

\noindent\textbf{Imposing uniqueness into CP decomposition through data augmentation.} We next employ a domain-knowledge led approach aimed at eliminating overlapping spectral content in CP factors and ensuring the uniqueness of the decomposition. This is achieved by augmenting the time dimension of the original tensor to increase the rank of factor matrices and obtain a clean and parsimonious frequency-channel separation. For clarity, we first demonstrate how this can be achieved on an idealized TFR, whereby for each data channel, only the magnitude of the most dominant frequency bin is set to one, and all other entries are zero, as shown in Figure \ref{fig.originalFakeTensor}. In other words, each of the three TFR channels contains a single frequency component, while the spectral domain is discretized into the corresponding three frequency bins. Upon applying the CP decomposition of rank-$3$ to this TFR, we obtain a sum of three rank-$1$ tensors, given by
\begin{equation}
   \label{eq.proof1}
   \ten{X} = \ten{A} + \ten{B} + \ten{C}
\end{equation}

These desired ``factor'' tensors \( \ten{A}, \ten{B}, \ten{C} \), illustrated in Figure \ref{fig.originalFakeTensor} (Top), are constructed as outer products of the TF factor matrices, $\mat{A}, \mat{B}, \mat{C}$, with canonical channel vectors $\mat{m}_1, \mat{m}_2, \mat{m}_3$, in the form
\begin{equation}
   \label{eq.proof2}
      \ten{A} = \mat{A} \circ \mat{m}_1 = \mat{A} \circ \begin{bmatrix} 1 \\ 0 \\ 0 \end{bmatrix}, \ 
      \ten{B} = \mat{B} \circ \mat{m}_2 = \mat{B} \circ \begin{bmatrix} 0 \\ 1 \\ 0 \end{bmatrix}, \ 
      \ten{C} = \mat{C} \circ \mat{m}_3 = \mat{C} \circ \begin{bmatrix} 0 \\ 0 \\ 1 \end{bmatrix}.
\end{equation}

The channel mixing vectors \( \mat{m}_1, \mat{m}_2, \mat{m}_3 \) form the columns of the channel mixing factor matrix \( \mat{M} \). Since we are considering an idealized TFR, each matrix \( \mat{A}, \mat{B}, \mat{C} \) is of size $3\times3$, corresponding to the three frequency bins and the three time points, and is defined to have only one non-zero row (individual frequency component), as shown in Figure \ref{fig.originalFakeTensor}, and given by
\begin{equation}
  \mat{A} =  \mat{f}_1 \circ \mat{t}_1 = 
  \begin{bmatrix} 1 \\ 0 \\ 0 \end{bmatrix} \begin{bmatrix} 1 & 1 & 1 \end{bmatrix}, \
  \mat{B} =\mat{f}_2 \circ \mat{t}_2 = 
  \begin{bmatrix} 0 \\ 1 \\ 0 \end{bmatrix} \begin{bmatrix} 1 & 1 & 1 \end{bmatrix}, \
  \mat{C} = \mat{f}_3 \circ \mat{t}_3 =
  \begin{bmatrix} 0 \\ 0 \\ 1 \end{bmatrix} \begin{bmatrix} 1 & 1 & 1 \end{bmatrix}. \
\label{eq.proof3}
\end{equation}

The vectors \( \mat{f}_1, \mat{f}_2, \mat{f}_3 \) and \( \mat{t}_1, \mat{t}_2, \mat{t}_3 \) constitute respectively the columns of the frequency factor matrix \( \mat{F} \) and time factor matrix \( \mat{T} \). Now, from Figure \ref{fig:tf_cp}, the resulting tensor, \( \ten{X} \), can be written as
\begin{equation}
   \label{eq.proof4}
   \ten{X} = \ten{I} \times_1^2 \mat{F} \times_1^2 \mat{T} \times_1^2 \mat{M},
\end{equation}
where $\ten{I}$ is a super-diagonal tensor with non-zero entries set to $1$. The domain knowledge of the TFR of the three sinusoids suggest a CP rank of $3$. Therefore, each of these factor matrices has three columns,
\begin{equation}
\label{eq:FTM}
\mat{F} = 
\begin{bmatrix}
1 & 0 & 0\\
0 & 1 & 0\\
0 & 0 & 1
\end{bmatrix}, \quad
\mat{T} = 
\begin{bmatrix}
1 & 1 & 1\\
1 & 1 & 1\\
1 & 1 & 1
\end{bmatrix}, \quad
\mat{M} = 
\begin{bmatrix}
1 & 0 & 0\\
0 & 1 & 0\\
0 & 0 & 1
\end{bmatrix},
\quad \text{where} \quad
\begin{aligned}
&\rank(\mat{F}) = 3,\\
&\rank(\mat{T}) = 1,\\
&\rank(\mat{M}) = 3.
\end{aligned}
\end{equation}

\begin{remark}
The Kruskal condition for the uniqueness of CP decomposition, in Equation (\ref{eq.uniq1}), states that
\begin{equation}
   \label{eq.proof5}
   \operatorname{rank}(\mat{F}) + \operatorname{rank}(\mat{T}) + \operatorname{rank}(\mat{M}) \geq 2R + 2 = 8,
\end{equation}
This inequality is not satisfied in our example as $\operatorname{rank}(\mat{F}) + \operatorname{rank}(\mat{T}) + \operatorname{rank}(\mat{M}) = 3+1+3=7 < 8$. Namely, in Equation (\ref{eq:FTM}), although the frequency and the channel mixing matrices, \( \mat{F} \) and \( \mat{M} \), are full-rank (their rank is $3$), the time factor matrix, \( \mat{T} \), has only a rank of $1$, which violates the Kruskal condition and prevents a unique CP decomposition. Therefore, to ensure identifiability, it is necessary for at least two columns of \( \mat{T} \) to be linearly independent to satisfy the uniqueness condition in Equation (\ref{eq.uniq1}).
\end{remark}

\noindent\textbf{Using domain knowledge to promote uniqueness of CP decomposition.} To resolve the uniqueness and consequently the explainability issues, we next employ domain knowledge to ensure the physical interpretability of the factor matrices. Since the frequency and channel modes, $\mat{F}$ and $\mat{M}$, are fixed by the acquisition set-up, they should remain unchanged unless new frequency content or channels are introduced. In contrast, we are free to augment the time mode, effectively modifying the time factor matrix, $\mat{T}$. This approach is illustrated in Figure \ref{fig.originalFakeTensor}, where additional time samples (designated in red) are appended to the original TFR tensor, resulting in a new, augmented factor matrix $\mat{\hat{T}}$, defined as
\begin{equation}
  \mat{\hat{T}} = [\mat{\hat{t}}_1 \hspace{.2cm} | \hspace{.2cm} \mat{\hat{t}}_2 \hspace{.2cm} |\hspace{.2cm} \mat{\hat{t}}_3]  
  = \begin{bmatrix}
  1 & 1 & 1\\ 
  1 & 1 & 1\\ 
  1 & 1 & 1\\ 
  \color{red}{1} & \color{red}{1} & \color{red}{0}\\ 
  \color{red}{1} & \color{red}{0} & \color{red}{0} 
  \end{bmatrix},
  \quad \text{where} \quad \operatorname{rank}(\mat{\hat{T}}) = 3.
  \label{eq.newT}
\end{equation}

\begin{remark}
The augmented time factor matrix \( \mat{\hat{T}} \) in Equation (\ref{eq.newT}) now has full column rank, that is, $\operatorname{rank}(\mat{\hat{T}}) = 3$, thereby satisfying the Kruskal condition in Equation (\ref{eq.proof5}), as now $\operatorname{rank}(\mat{F}) + \operatorname{rank}(\mat{M}) + \operatorname{rank}(\mat{\hat{T}}) = 9 > 8$. The CP decomposition of such an ``augmented'' tensor is not only guaranteed to yield a unique solution, but it also allows for perfect separation of the frequencies in each CP component.
\end{remark}

This gives us the opportunity to revisit the real-life (non-ideal) TFR tensor example in Figure \ref{fig:tf_cp} (Bottom Right). The strategy of augmenting the temporal mode increases the rank of the temporal factor matrix, so as to meet the Kruskal uniqueness condition in Equation (\ref{eq.proof5}), indeed yields the desired separation of original signal channels in the CP components. In this way, such simple yet insightful strategy preserves the explainability of the frequency and channel mode components while ensuring mathematical identifiability through a domain-informed augmentation of the time mode.

\begin{remark}
    The ALS algorithm for the CP decomposition yields a unique decomposition if the Kruskal uniqueness condition is satisfied. The use of the available domain knowledge gives us the freedom to make amendments to the original multi-dimensional time-frequency array, without loss of generality, and in this way helps enforcing the ALS to provide a unique CP decomposition. Despite the fact that it is impossible to predict the resulting factor matrices for an unknown data set, the CP decomposition will still be unique up to standard scalings, and is given by
    \begin{equation}
          \ten{X} = \sum_{r=1}^R (a^{(1)}_r \mathbf{a}^{(1)}_r) \circ \cdots \circ (a^{(N)}_r \mathbf{a}^{(N)}_r)
    \end{equation}
    
    Here, the product of all coefficients $\allowbreak a^{(1)}_r, a^{(2)}_r, \dots, a^{(N)}_r$ is equal to unity for any $r$. We refer the reader to \cite{6544287} for more information on the Alternating Least Squares (ALS) algorithm for CP decomposition.
\end{remark}

\subsection*{Illuminating TN Ranks through RGB Color Image Representation}
\label{sec:color}
\begin{figure*}
    \centering
    \includegraphics[width=1\columnwidth]{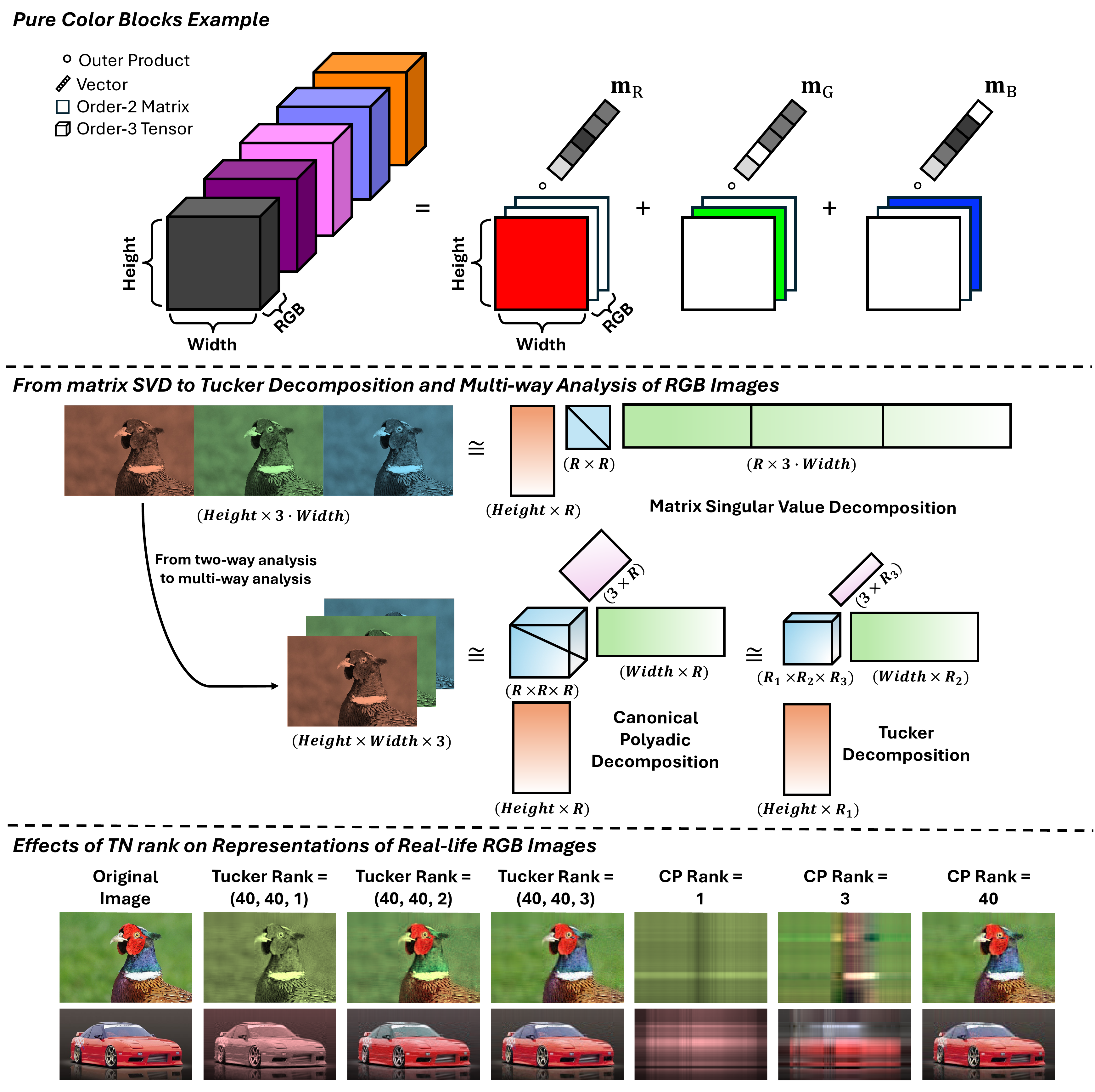}
    
    \caption{
        The concept of minimal TN rank illuminated through RGB representation.
        \textbf{Top:} Block term decomposition of five pure-color RGB images stacked together into a fourth-order tensor. This tensor is factorized into three base color components (red, green, blue) and a color mixing matrix, $\mat{M} = [\mat{m}_R | \mat{m}_G | \mat{m}_B]$, so that each color can be perfectly reconstructed via a linear combination of these three rank-$1$ color bases.
        \textbf{Middle:} CP decomposition enables decomposition of higher-order data into rank-$1$ components, while Tucker decomposition generalises the CP decomposition and has a vector-valued rank \cite{HOSVD}.
        \textbf{Bottom:} Compression of real-world RGB images using the CP and Tucker decompositions of varying ranks. Tucker decompositon exhibits mode-specific rank control due to its vector nature of ranks. On the contrary, the ``scalar'' CP rank has limited control over individual modes, whereby an increase in rank improves clarity but lacks fine-grained flexibility.
    }
    \label{fig:color-ensemble}
\end{figure*}

Similar to the context of TFR tensors, we next show that domain knowledge can serve to provide an informed way to  determine the TN ranks in RGB image tensors. To this end, consider five RGB images of different pure colors, $\{\ten{X}_n\}_{n=1}^5 \in \mathbb{R}^{\text{Height} \times \text{Width} \times \text{RGB channels}}$, stacked together to constitute an order-$4$ tensor \(\ten{\hat{X}} \in \mathbb{R}^{\text{Height} \times \text{Width} \times \text{RGB Channels} \times 5}\), as illustrated in Figure~\ref{fig:color-ensemble} (Top). Colors in digital images are represented by the combination of pixel values (from $0$ to $255$) of the red, green, and blue (RGB) color channels. This domain knowledge suggests that any color can be represented as a linear combination of the three base colors, red, green, and blue. In other words, there exists an optimal low-rank TN representation of TN rank 3, that is, the number of base colors (red, green, blue), which allows for perfect reconstruction of the original tensor. Since each pure color, $\mathcal{X}_n$, is an order-$3$ tensor, the order-$4$ tensor of stacked pure colors, $\ten{\hat{X}}$, can be decomposed through a block-term decomposition into a smaller ``core'' tensor of the same order, $\ten{Z} \in \mathbb{R}^{\text{Height} \times \text{Width} \times \text{RGB Channels} \times 3}$, and a color mixing matrix $\mat{M} \in \mathbb{R}^{5 \times 3}$, as shown in Figure~\ref{fig:color-ensemble} (Top). Then, the three order-$3$ frontal slices of $\ten{Z}$ represent pure color images corresponding to the three base colors.

The set-up in Figure \ref{fig:color-ensemble} (Top) can be formalized as
\begin{equation}
    \mathcal{Z}(\text{Height},\text{Width},\text{RGB Channel},\text{Base Color}) \;=\;
    \begin{cases}
      1, & \text{RGB Channel} = \text{Base Color},\\[4pt]
      0, & \text{RGB Channel} \neq \text{Base Color}.
    \end{cases}
\end{equation}

with the corresponding color mixing vectors as

\begin{equation}
\begin{aligned}
\mat{m}_R &= \begin{bmatrix} 64\\ 128\\ 255\\ 128  \\ 128 \end{bmatrix} \quad
\mat{m}_G = \begin{bmatrix} 64\\ 0\\ 128  \\ 128\\ 128 \end{bmatrix} \quad
\mat{m}_B = \begin{bmatrix} 64\\ 128  \\ 255\\ 255\\ 0  \end{bmatrix} \\
\end{aligned}
\end{equation}

Then, the tensor, $\ten{\hat{X}}$, containing the five pure colors in Figure \ref{fig:color-ensemble} (Top) can be expressed as
\begin{equation}
    \ten{\hat{X}} = \ten{Z} \times_4^2 \mat{M}\\[1ex]
    = 
    \underbrace{\ten{Z}(:,:,:,1)}_{\text{Red}} \circ \mat{m}_R
    + \underbrace{\ten{Z}(:,:,:,2)}_{\text{Green}} \circ \mat{m}_G
    + \underbrace{\ten{Z}(:,:,:,3)}_{\text{Blue}} \circ \mat{m}_B
\end{equation}

\begin{remark}
The domain knowledge within the RGB representation allows for a perfect reconstruction of the original ``pure colors'' tensor through a low-rank representation, whereby the $5$ different pure color images in the tensor $\ten{\hat{X}}$ can be represented by different scalings of the three base RGB color images. This observation extends to any number of colors, illustrates how TN rank directly follows from physical properties of the data, and reinforces the physical intuition behind low-rank TN structures.
\end{remark}

\subsection*{Tucker Decompositon and Vector-Valued TN Rank} The Tucker decomposition generalizes the CP decomposition by decomposing a tensor into a smaller-scale dense core tensor, $\ten{G}$, which is connected via tensor contractions with factor matrices, along each of its modes, as shown in Figure \ref{fig:color-ensemble} (Middle) and Figure \ref{fig:Example1} (Bottom Left). For an order-$3$ tensor, the Tucker decomposition takes the form
\begin{equation}
\label{eq:Tucker}
    \ten{X}(i_1, i_2, i_3) = \ten{G} \times_1^2 \mat{A} \times_1^2 \mat{B} \times_1^2 \mat{C} = \sum_{r_1=1}^{R_1} \sum_{r_2=1}^{R_2} \sum_{r_3=1}^{R_3} \ten{G}(r_1, r_2, r_3) \mat{A}(i_1, r_1) \mat{B}(i_2, r_2) \mat{C}(i_3, r_3)
\end{equation}
where $\mat{A}, \mat{B},$ and $\mat{C}$ are the factor matrices. Observe from Equation (\ref{eq.CPD1}), Equation (\ref{eq:Tucker}), and Figure \ref{fig:Example1} (Bottom Left) that the CP decomposition is a special case of the Tucker decomposition, whereby the core tensor $\mathcal{G}$ is sparse and \textit{super-diagonal}, i.e., $g(r_1, r_2, r_3) = 0 \text{ unless } r_1 = r_2 = r_3$ in the order-$3$ case. This subtle difference has a significant impact on the corresponding ranks. 

\begin{remark}
Unlike in the CP decomposition, where the rank is a scalar, the TN rank of Tucker decomposition is a vector, $[ R_1, R_2, \ldots, R_N]$, where $N$ is the order (number of modes) in the original tensor, and $R_n, \ n=1,\ldots,N,$ is the rank for the $n$-th mode. Similarly, more complex TN structures, such as the Tensor Train (TT) \cite{TT_decomposition} and Fully Connected Tensor Network (FCTN) \cite{Zheng_Huang_Zhao_Zhao_Jiang_2021}, have a vector rank.
\end{remark}

\subsection*{Example 2: TN ranks and Model Expressivity} To demonstrate the trade-offs between model expressivity and low-rankness, we consider the real-world ``Bird'' and ``Car'' RGB images given in Figure \ref{fig:color-ensemble} (Bottom), which are represented by an order-$3$ tensor of the form $\ten{X} \in \mathbb{R}^{\text{Height} \times \text{Width} \times \text{RGB Channels}}$. Note that a lower rank also means fewer parameters required by the model. Figure \ref{fig:color-ensemble} (Bottom) demonstrates the compressed images for the ``Bird'' and ``Car'' images, when using the CP and Tucker decompositions with different ranks. From the pure color images example in Figure \ref{fig:color-ensemble} (Top), we know that a TN rank of $3$ in the RGB channel mode is required to preserve the color information in the RGB images. Observe that, the ``Bird'' image contains prominent pure color parts of all three base colors, indicating that it requires at least a TN rank of $3$ to reconstruct the original image well. When a TN rank of $1$ is used in the RGB channel mode of the Tucker decomposition, the reconstructed ``Bird'' image only contains different shades of green, as the Tucker decomposition is obtained by minimizing the Frobenius norm between the original tensor and the reconstructed one, given a vector of ranks. Furthermore, when a TN rank of $2$ is used in the RGB channel mode of the Tucker decomposition, observe that the brown and red components are correctly reconstructed, but the blue component of the ``Bird'' is now incorrectly shown as green, effectively indicating that only two base colors are preserved. When a TN rank of $3$ is used in the RGB channel mode of the Tucker decomposition, the color information of the image is correctly preserved. On the other hand, the ``Car'' image has less variation in its color, as the most prominent features are the red car body, the blue-ish car windows, and the black-ish background. Therefore, as shown in Figure \ref{fig:color-ensemble} a TN rank of $2$ in the RGB channel mode of the Tucker decomposition is already sufficient to preserve most of the color information in the ``Car'' image, whereas a TN rank of $3$ in the RGB channel achieves a perfect color reconstruction.

\begin{remark}
Unlike pure color images in Figure \ref{fig:color-ensemble} (Top), real-world RGB images also exhibit variations in the vertical and horizontal pixel values which contain rich information regarding the photographed object. Therefore, to avoid a blurry reconstruction of the original images, sufficient TN ranks are required in the spatial modes. Since the rank in the CP decomposition is a scalar shared across all tensor modes, if we set the CP rank to $3$ as shown in Figure \ref{fig:color-ensemble} (Bottom), although the color information can be preserved, the reconstructed image is very blurry. When the CP rank is increased to $40$, we obtain much clearer images, but this also leads to a unnecessary high CP rank in the RGB color mode. This demonstrates that the CP decomposition provides a less intricate control over the reconstruction quality compared to the Tucker decomposition, as its TN rank is a scalar and not a vector as in the Tucker decomposition.
\end{remark}

\subsection*{Connecting the unfolded matrix ranks and TN ranks with TN graphs}
\begin{figure*}
\includegraphics[width=1\columnwidth]{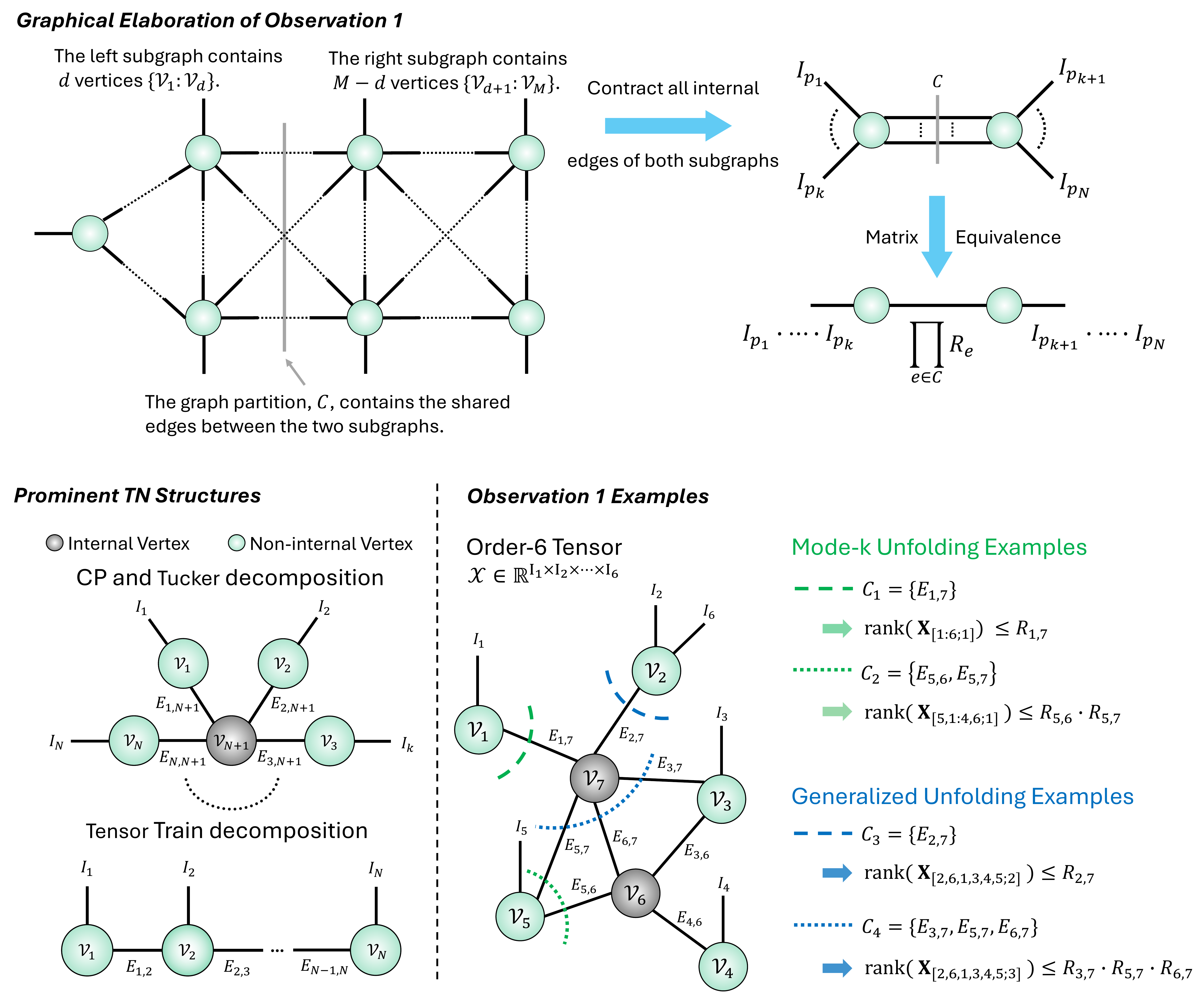}
\caption{Connecting TN ranks and matrix ranks through TN diagrams. \textbf{Top:} Graphical elaboration of Observation \ref{assertion2}. \textbf{Bottom Left:} Prominent TN structures such as the CP decomposition, Tucker decomposition, and Tensor Train decomposition. The CP decomposition exhibits the same TN topology as the Tucker decomposition, but has a super-diagonal core tensor $\mathcal{V}_{N+1}$. \textbf{Bottom Right:} Examples of using Observation \ref{assertion2} and the graphical interpretation of the unfolding ranks on TN graphs. Observation \ref{assertion2} allows us to find the unfolding rank bounds by directly \textit{partitioning} the TN graph and thus completely avoiding mathematical equations and index manipulations. The unfolding matrix rank bounds are obtained by simply multiplying the rank values of the edges in each TN graph partition.}
\label{fig:Example1}
\end{figure*}

We have seen that the concept of TN rank does not have a unique definition across different TN structures, as e.g., indicated by the edge values (connectivities) between vertices in the TN graphs shown in Figure \ref{fig:Example1} (Bottom Left) \cite{orus2014practical}. However, since tensor algebra is a higher-order extension of the linear algebra of matrices, the TN ranks admit a natural treatment through a connection with the matrix ranks of the unfolded tensor \cite{domanov2021computation,Zheng_Huang_Zhao_Zhao_Jiang_2021, HOSVD}. Nonetheless, deriving this relation for complex TN structures usually involves cumbersome mathematical equations and manipulation of multi-indexed tensors, which may be overwhelming for general practitioners. At the same time, the relation between TN ranks and matrix ranks for different tensor networks is of great importance, as it allows the examination of different TN properties through intuitive links with their matrix counterparts \cite{TT_decomposition, CALVI2021107862}.  We next explore these links through a diagrammatic procedure, stated in Observation \ref{assertion2}, which bypasses the tensor maths and provides an intuitive understanding of how TN ranks relate to matrix ranks for any TN structure. The corresponding self-understood graphical elaboration, suitable for a generally knowledgeable reader, is provided in Figure \ref{fig:Example1} (Top). For completeness, the full mathematical description is provided in the Supplementary Material. Figure \ref{fig:Example1} (Bottom Right) also offers four intuitive and self-explanatory examples of the usage of this graphical approach. 

\subsection*{TN ranks and Graphical Diagrams of Tensor Networks}
A generalized TN is composed of a set of smaller-sized tensors which are contracted together to represent the original order-$N$ tensor, $\mathcal{X}\in \mathbb{R}^{I_1 \times I_2 \times \cdots \times I_N}$. As shown in Figure \ref{fig:Example1} (Bottom Left), the CP and Tucker decompositions can both be represented through a TN diagram, via vertices and edges, while their corresponding ranks are reflected through the edges. The graphical diagram of a TN is an undirected graph, $G = (V, E)$, where $V = (\mathcal{V}_1, \mathcal{V}_2, \ldots, \mathcal{V}_M)$, is the set containing the $M$ smaller-sized ``core'' tensors obtained through TN decompositions. The set of connections between the vertices are called the closed edges, $E = (E_{1,2}, E_{1,3}, \ldots, E_{M-1,M}): i<j$, where $i$ and $j$ are the indices of the connected vertices. We also denote the complete set of closed edges connected to a vertex $\mathcal{V}_k$ by $E_k \subseteq E$. Each closed edge has an assigned (edge connectivity) value, the so called TN rank $R_{i,j} ( R_{i,j} \in \mathbb{Z}^+, i < j)$, which represents the mode size of the connected vertices. When a TN rank is equal to unity, i.e., $R_{i,j} = 1$, that edge can be dropped to simplify the graph. Therefore, $E_{i,j} = \{R_{i,j}\}$ when $R_{i,j}>1$, and $E_{i,j} = \emptyset$ when $R_{i,j}=1$. An open edge of a vertex is associated with an ``external'' mode of the original tensor that is not included in $E$. A graph is called a connected graph if there exists a path between every pair of vertices or if it contains only one vertex. Figure \ref{fig:Example1} (Bottom Left) also shows some popular TN formats and the difference between an internal vertex and a non-internal vertex. A TN graph partition, $C \subset E$, disconnects parts of a TN by cutting the edges within a TN, and is defined by the set of closed edges it cuts through.

\begin{thm}[\textbf{General Unfolding Rank Bounds via TN graphs}]
Consider an order-$N$ tensor, $\mathcal{X} \in \mathbb{R}^{I_1 \times I_2 \times \dots \times I_N}$, which has a tensor network representation, $G = (V, E)$, with $M$ vertices. Given a mode permutation vector, $\mathbf{p} = (p_1, p_2, \ldots, p_N)$, there exist two integers, $k$ and $d$, such that $1 \leq k \leq N, 1 \leq d \leq M$, and a non-empty TN graph partition, $C$, which separates the tensor network, $G$, into two disjoint connected subgraphs, containing $\{ \mathcal{V}_{p_i} \}_{i=1}^{d}$ vertices associated with modes $\{ p_1, p_2, \dots, p_k \}$ and $\{ \mathcal{V}_{p_j} \}_{j=d+1}^{M}$ vertices associated with modes $\{ p_{k+1}, p_{k+2}, \dots, p_N \}$. Then, the rank of the generalized matrix unfolding of $\mathcal{X}$, denoted by $\mathbf{X}_{[\mathbf{p}, k]}$, satisfies
\begin{equation}
\operatorname{rank}\left( \mathbf{X}_{[\mathbf{p}, k]} \right) \leq \prod_{e \in C} R_e.
\end{equation}
where $R_e$ are the rank (edge connectivity) values of the edges that the graph partition $C$ cuts through.

\label{assertion2}
\end{thm}

\begin{remark}
Observation \ref{assertion2} allows for the analysis of the TN ranks in an intuitive and graphical way, without resorting to multi-linear algebra. This is achieved by connecting TN ranks to the corresponding matrix rank of the unfolded tensor, and can be flexibly applied to TN structures of any order. 
\end{remark}

Figure \ref{fig:Example1} (Bottom Right) provides four examples of the applicability of Observation \ref{assertion2}. Assume, for example, an order-$6$ tensor. Our aim is to represent it compactly with a TN of $5$ non-internal vertices and $2$ internal vertices. To examine the suitability of such a TN structure in specific applications, we examine its expressivity in different modes and verify the results via domain-meaningful priors. Upon applying Observation \ref{assertion2}, we can see that the mode-$1$ and mode-$5$ unfolding ranks of the TN in Figure \ref{fig:Example1} (Bottom Right) are respectively upper bounded by $R_{1,7}$ and $R_{5,6}\cdot R_{5,7}$. Assuming, for simplicity, all rank values to be $R$, such a TN structure can allow more variability in its $5$th mode compared to its $1$st mode in relation to the rest of the modes, as the rank upper bound for the mode-$5$ unfolding is $R^2$ while the rank upper bound for the mode-$1$ unfolding is only $R$. This allows us to incorporate domain-knowledge about the data, whereby the $5$th mode could contain more variability than the $1$st mode. Upon applying Observation \ref{assertion2} to the TN structure in Figure \ref{fig:Example1} (Bottom Right), observe that its generalized unfolding ranks are also upper bounded by the product of the rank values in the corresponding TN graph partitions.

\begin{remark}
For the TN structure in Figure \ref{fig:Example1} (Bottom Right), it is not possible to establish an upper bound for the rank of its mode-$2$ or mode-$6$ unfoldings using Observation \ref{assertion2}. This is because mode-$2$ and mode-$6$ are associated with the same vertex, $\mathcal{V}_2$. Consequently, any graph partition will place mode-$2$ and mode-$6$ in the same subgraph, making it impossible to isolate either mode from the other as required by Observation \ref{assertion2}. This implies that such a TN structure does not explicitly upper bound the mode-2 and mode-6 unfolding ranks. This inherent flexibility can be advantageous in applications where preserving maximal degrees of correlation of certain modes is desirable, such as when modeling complex inter-modal interactions that should not be artificially bounded by the TN topology.
\end{remark}

\subsection*{Example 3: TN Ranks and TN Design}

Understanding the unfolded matrix ranks of TNs gives useful insight when designing appropriate TN structures by using domain-specific priors. This is next demonstrated through the example of tailoring a TN structure for compressing an order-$4$ RGB ``News'' video data tensor, $\mathcal{G}\in \mathbb{R}^{144\times 176 \times 3 \times 50}$, used in \cite{Zheng_Huang_Zhao_Zhao_Jiang_2021}. The four modes of the original tensor correspond, respectively, to the height ($144$ pixels), width ($176$ pixels), RGB channels ($3$ channels), and frame number ($50$ frames). Our goal is to achieve as much space saving as possible through tensor compression while keeping the approximation error low.

\begin{figure}
    \centerline{\includegraphics[width=1\columnwidth]{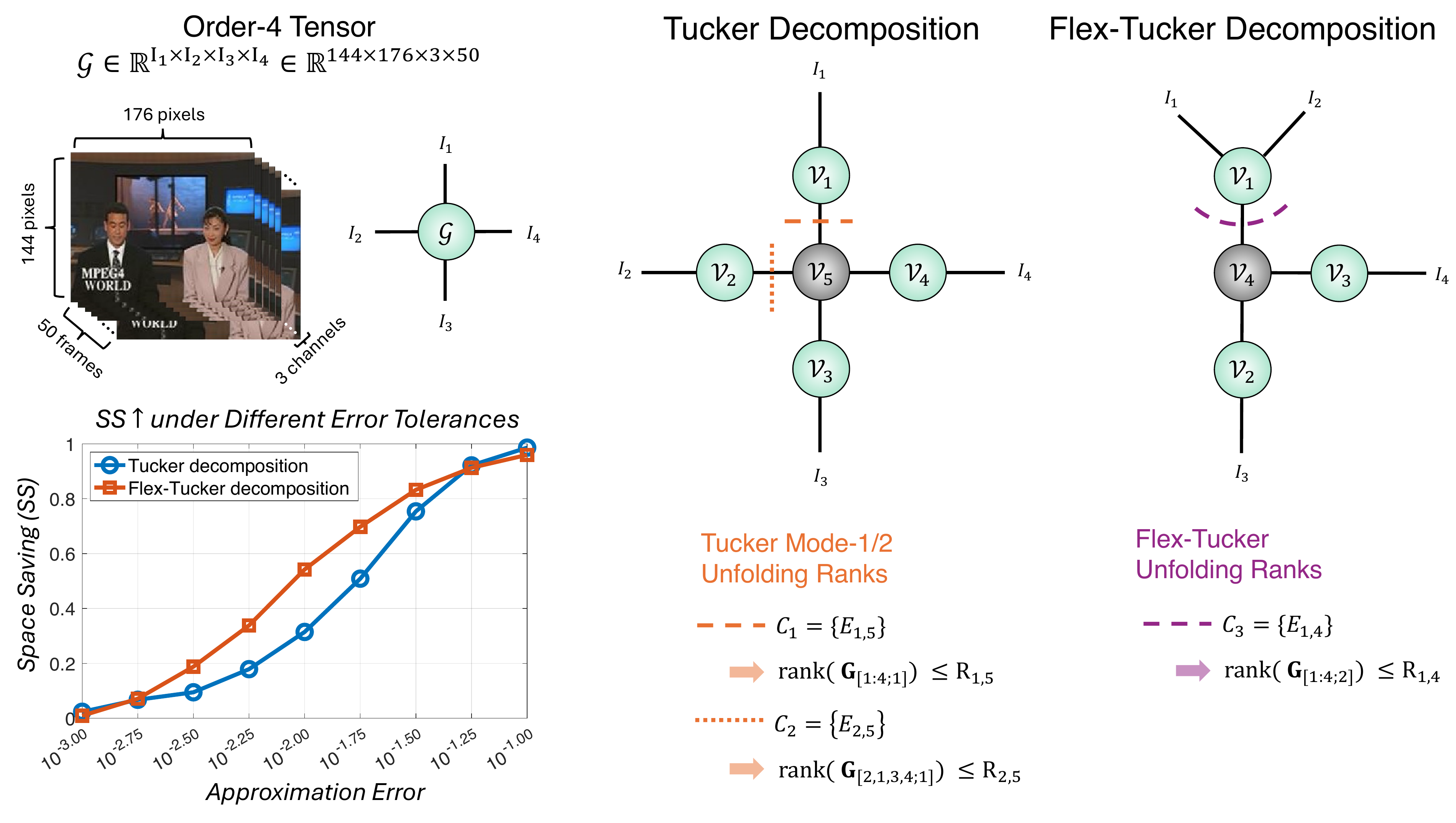}}
    \caption{The TN graphs of the Tucker decomposition and the Flex-Tucker decomposition for the order-$4$ RGB ``News'' video data tensor, $\mathcal{G}\in \mathbb{R}^{144\times 176 \times 3 \times 50}$. Frames of the video tensor are shown at the top left. Upon applying Observation \ref{assertion2}, we see that the Tucker decomposition upper bounds its mode-$1$, mode-$2$, mode-$3$, and mode-$4$ unfolding ranks through respective graph partitions, while the Flex-Tucker decomposition does not upper bound individual mode-$1$ and mode-$2$ unfolding ranks. The bottom left plot shows the space saving $\left( \text{SS} = \frac{\text{Original Size}-\text{Compressed Size}}{\text{Original Size}}  \right)$ of the Tucker decomposition and the Flex-Tucker decomposition under different approximation error tolerances. Unlike the standard Tucker decomposition, the Flex-Tucker decomposition has a vertex $\mathcal{V}_1$, which contains both external mode-$1$ ($I_1$) and mode-$2$ ($I_2$), and is thus able to compress the original tensor more efficiently than the Tucker decomposition under most approximation errors by preserving the high correlation between the height (mode-$1$) and width (mode-$2$) modes in the RGB video tensor.}
    \label{fig:Example2}
\end{figure}

A standard approach for compressing order-$4$ RGB video tensors is the Tucker decomposition, whose TN structure is shown in Figure \ref{fig:Example2}. According to Observation~\ref{assertion2}, the Tucker format imposes an upper bound on the unfolding rank of every mode, as each factor matrix vertex is connected to exactly one external mode. However, we have prior knowledge about this “News” RGB video tensor, as the spatial modes (height and width) have much larger sizes ($144$ and $176$) than the RGB and frame modes (of size $3$ and $50$) and contain complex spatial information. This imbalance suggests that an optimal TN structure should be flexible enough to capture the rich information in the spatial modes, while achieving compression by constraining the unfolded matrix ranks associated with the RGB and frame modes. 

This objective is achieved by employing the available domain knowledge to define a new TN structure, termed the \textit{Flex-Tucker decomposition}, as shown in Figure \ref{fig:Example2}. The Flex-Tucker TN structure allows for the external large-sized height and width modes to share a common vertex, $\mathcal{V}_1$ to preserve the intricate spatial correlations, which is perfectly physically meaningful for RGB color videos. While this also means that it is no longer possible to establish an upper bound for the individual mode-$1$ or mode-$2$ unfolding ranks using Observation \ref{assertion2}, the use of Observation \ref{assertion2} still yields the same rank upper bounds on the mode-$3$ (RGB) and mode-$4$ (frame) unfoldings as the Tucker decomposition. In other words, the Flex-Tucker decomposition allows us to bypass the upper bounds on the ranks of the mode-$1$ and mode-$2$ unfoldings, and thus preserve the strong spatial correlations between the height and width modes while enabling efficient tensor compression.

The virtues of the additional flexibility of TN design established through domain knowledge and TN rank analysis are demonstrated on the task of compressing the RGB ``News'' video tensor, $\mathcal{G}$, using both the standard Tucker and the Flex-Tucker decompositions. Figure \ref{fig:Example2} illustrates the achieved space saving (SS), defined as $ \text{SS} = \frac{\text{Original Size}- \text{Compressed Size}}{\text{Original Size}}$,  against the approximation error, calculated as $\frac{\| \text{Original Tensor} - \text{Approximated Tensor} \|_F}{\| \text{Original Tensor} \|_F}$. The results clearly show that under most given approximation error tolerances, the Flex-Tucker decomposition achieves superior space savings than the Tucker decomposition. This confirms that by leveraging the unfolding rank properties of a TN through domain knowledge, we can design structures that are better suited to the specific characteristics of the data.

\section*{What we have learned}
Tensor network (TN) ranks underpin TN decompositions, yet they lack a universal interpretation across different TN formats and remain less well understood than the matrix rank. To this end, we have systematically addressed the often elusive and confusing notion of TN ranks through the integration of domain knowledge in order to promote model explainability, ensure uniqueness, and enhance design flexibility. This perspective is supported by illustrative examples in practical settings, ranging from time-frequency signal representations to RGB image and video tensors, which demonstrate how the choice of TN ranks directly impacts these desirable properties. Furthermore, we have employed a unifying graphical framework which allows practitioners to analyze upper bounds on matrix unfolding ranks based solely on the topology of TN graphs. This diagrammatic approach bypasses the need for intricate and cumbersome manipulations in multilinear algebra. The so introduced ability to ``read off" rank bounds from TN graphs also empowers users to design TNs informed by rank theory, rather than relying on trial-and-error tuning. Overall, this Lecture Note has revisited the notion of tensor network (TN) ranks via a principled approach which rests upon the use of domain knowledge, which is particularly useful in settings where ranks are linked to physical or semantic properties of the data. It is our hope that such a perspective, grounded in first principles and domain knowledge,  will equip  practitioners with more intuition and flexibility in their design, while demystifying tensor ranks for educational purposes.

%
%
%
%
%
\bibliographystyle{IEEEtran}
\bibliography{reference}

\section*{Supplementary Material}

\begin{figure*}
    \centering
    \centerline{\includegraphics[width=1\columnwidth]{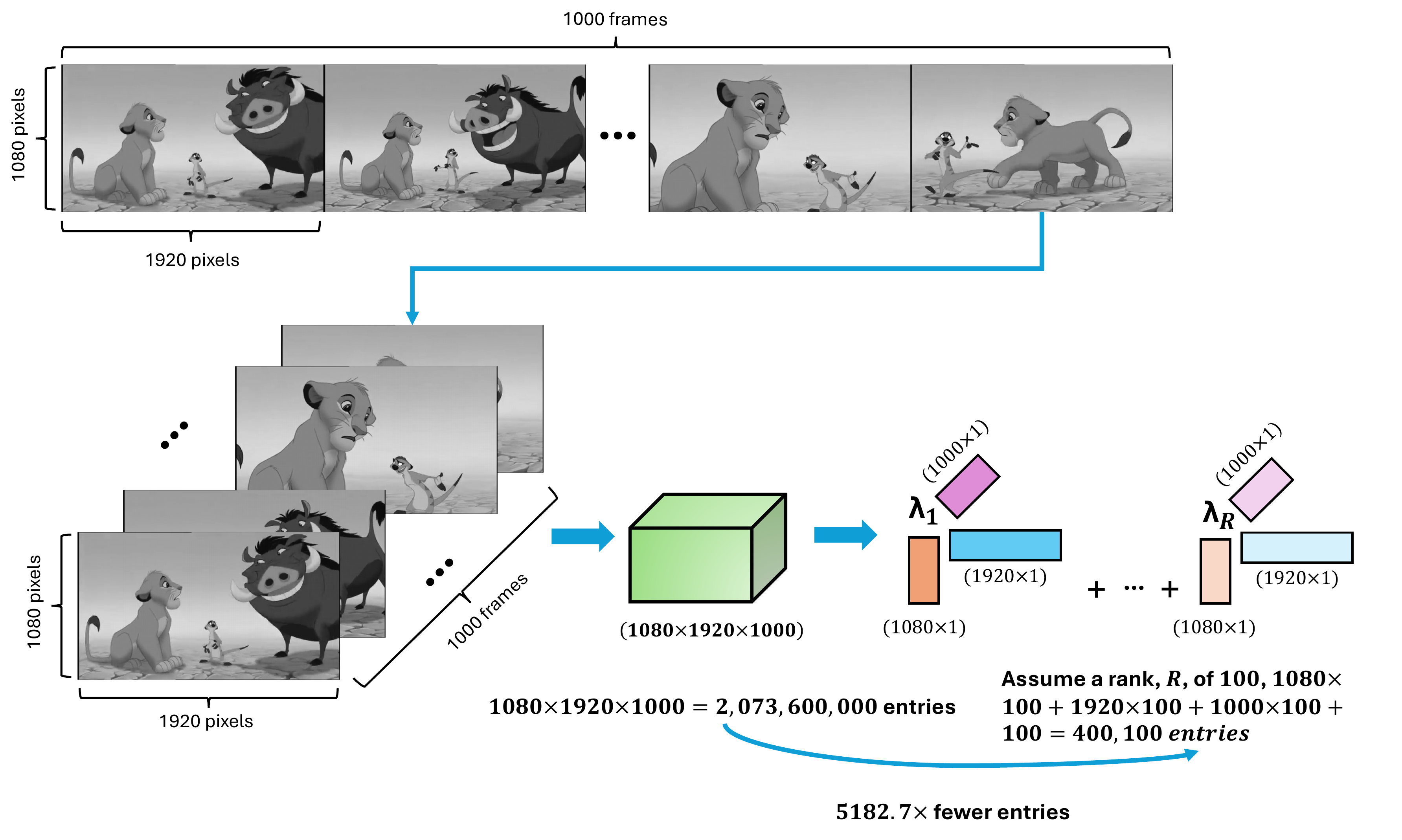}}
    \caption{Super-compression through the tensor Canonical Polyadic decompositions. \textbf{Top:} The $1000$ black and white frames from the movie, ``Lion King''. \textbf{Bottom:} Stacking the frames into a cube yields an order-$3$ tensor of dimensionality, $1080 \times 1920 \times 1000$, with $2,073,600,000$ entries. The CP decomposition with rank $100$, when applied to this order-$3$ tensor, only needs $400,100$ entries, which is $5182.7\times$ less than in the original tensor.}
    \label{fig:lion_king}
\end{figure*}

\section*{Super-compression enabled by tensor decompositions}
\begin{figure*}
    \centering
    \centerline{\includegraphics[width=0.6\columnwidth]{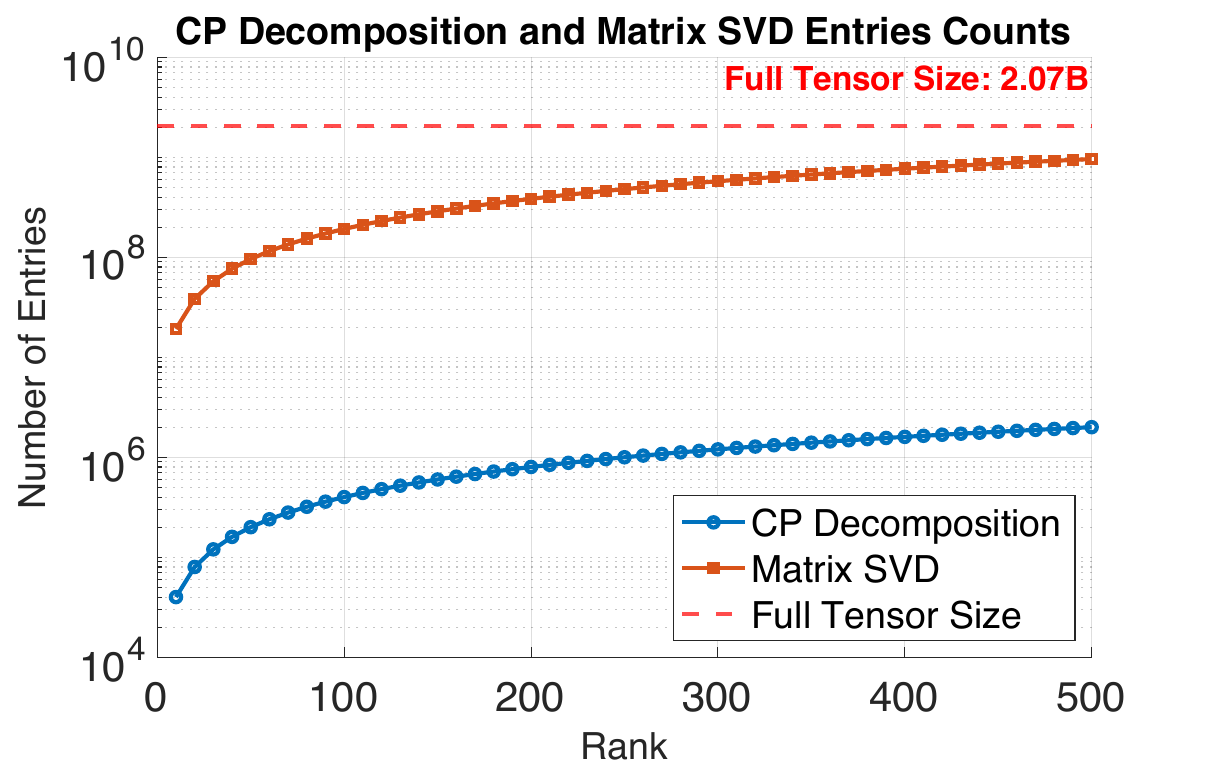}}
    \caption{Comparison for the number of entries after compressing the ``Lion King'' tensor (of size $1080 \times 1920 \times 1000$) in Figure \ref{fig:lion_king} with CP decomposition and its flattened matrix version (of size $1080 \times 1,920,000$) with matrix SVD. CP decomposition consistently requires considerably fewer parameters (entries) compared to matrix SVD under the same rank. The number of entries in the original tensor is $2,073,600,000$.}
    \label{fig:params}
\end{figure*}

The storage complexity of a tensor, in its original format, scales exponentially with the number of dimensions (modes). On the other hand, tensor decompositions can achieve super-compression by exploiting the low-rank structure of the original tensor as demonstrated in Figure 1 (Bottom) in the main text. Figure \ref{fig:lion_king} shows how by stacking the $1000$ black and white frames ($1080 \times 1920$ pixels) from the movie, ``Lion King'', into an order-$3$ tensor of dimensionality, $1080 \times 1920 \times 1000$, with $2,073,600,000$ entries, the CP decomposition with rank $100$ only needs $400,100$ entries, which is $5182.7\times$ less than the original tensor format. Note that a matrix singular value decomposition (SVD) of the original matrix, even with rank-$1$, would require $1 \times 1080 + 1\times 1000 \times 1920 = 1,921,080$ entries, which is $4.8\times$ more than the CP decomposed format of rank $100$. Also, note that increasing the rank of the CP decomposition by $1$ increases the number of entries by $1080+1920+1000=4,000$ entries, whereas increasing the rank of the SVD by $1$ increases the number of entries by $1080+1000\times1920=1,921,080$ entries. Figure \ref{fig:params} and Table \ref{fig:params} demonstrate how the total number of parameters in both CP decomposition and SVD changes as their ranks vary.

\begin{table}[ht]
\centering
\caption{Number of entries and the corresponding compression rates of CP decomposition and matrix SVD at different ranks when compressing the ``Lion King'' tensor of size $1080 \times 1920 \times 1000$ in Figure~\ref{fig:lion_king} (unfolded to size $1080 \times 1{,}920{,}000$ for matrix SVD). Compression rate is defined as $\frac{\text{Number of parameters in the original tensor}}{\text{Number of parameters in the compressed representation}}$.}
\label{tab:cp_svd_compression}
\begin{tabular}{|c|cc|cc|}
\hline
\multirow{2}{*}{\textbf{Rank}} & \multicolumn{2}{c|}{\textbf{CP Decomposition}} & \multicolumn{2}{c|}{\textbf{ Matrix SVD}} \\ \cline{2-5}
 & \textbf{No. of Entries} & \textbf{Compression Rate} & \textbf{No. of Entries} & \textbf{Compression Rate} \\ 
\hline
10  & 40{,}010       & $51{,}827\times$  & 19{,}210{,}800  & $107.9\times$ \\
50  & 200{,}050      & $10{,}365\times$  & 96{,}054{,}000  & $21.6\times$ \\
100 & 400{,}100      & $5{,}183\times$   & 192{,}108{,}000 & $10.8\times$ \\
500 & 2{,}000{,}500  & $1{,}037\times$   & 960{,}540{,}000 & $2.2\times$ \\
\hline
\end{tabular}
\end{table}

\section*{Mathematical Justification of Observation 1}

\begin{definition}
    Tensor contractions between two tensors are only allowed between modes of common sizes. For example, to contract the order-$N$ tensor, $\mathcal{X}\in \mathbb{R}^{I_1 \times I_2 \times \cdots \times I_N}$, with the order-$M$ tensor, $\mathcal{Y}\in \mathbb{R}^{J_1 \times J_2 \times \cdots \times J_M}$, we first let the two permutation vectors, $\mathbf{p^x}$ and $\mathbf{p^y}$, have their first $k \left( 1 \leq k \leq \operatorname{min}(N,M)\right)$ elements correspond to the $k$ common-sized modes in $\mathcal{X}$ and $\mathcal{Y}$, while the other entries follow their original ascending order. A tensor contraction is then given by
\begin{equation}
\begin{split}
    \mathcal{Q}&= \mathcal{X} \times^{{\mathbf{p^y}(1), \mathbf{p^y}(2), \ldots, \mathbf{p^y}(k)}}_{\mathbf{p^x}(1), \mathbf{p^x}(2), \ldots, \mathbf{p^x}(k)} \mathcal{Y} = \text{Fold}_{[(1,2,\ldots, M+N-2k);M-k]} (\mathbf{X}^{\mathsf{T}}_{[\mathbf{p^x};k]}  \mathbf{Y}_{[\mathbf{p^y};k]}) \\
    &\in \mathbb{R}^{I_{\mathbf{p^x}(k+1)} \times I_{\mathbf{p^x}(k+2)} \times \cdots \times I_{\mathbf{p^x}(N)} \times J_{\mathbf{p^y}(k+1)} \times J_{\mathbf{p^y}(k+1)} \times \cdots \times J_{\mathbf{p^y}(M)}}
    \label{equ:tensor_contraction2}
\end{split}
\end{equation}

\end{definition}

For clarity, we first introduce a simplified notation of tensor contractions, which is equivalent to Equation (\ref{equ:tensor_contraction2}) and is given by
\begin{equation}
    \mathcal{X} \times_{\{ E_{shared} \}} \mathcal{Y} = \mathcal{X} \times^{{\mathbf{p^y}(1), \mathbf{p^y}(2), \ldots, \mathbf{p^y}(k)}}_{\mathbf{p^x}(1), \mathbf{p^x}(2), \ldots, \mathbf{p^x}(k)} \mathcal{Y}
\end{equation}
where the set $\{E_{shared}\}$ comprises the shared edges between tensor $\mathcal{X}$ and tensor $\mathcal{Y}$, and $\| E_{shared} \|_0 = k$, where $\| \cdot \|_0$ is the $l_0$-norm. Denote by $C$ the graph partition which separates a TN into two connected TN subgraphs, $G_{sub,1}$ and $G_{sub,2}$. Let the connected TN subgraph containing $\{ \mathcal{V}_{p_i} \}_{i=1}^{d}$ be $G_{sub,1}$ and the connected subgraph containing $\{ \mathcal{V}_{p_j} \}_{j=d+1}^{M}$ be $G_{sub,2}$. Let $\| C\|_0=L$. We can obtain an order-$(L+k)$ tensor $\mathcal{Z} \in \mathbb{R}^{ I_{p_1} \times I_{p_2} \times \cdots \times I_{p_k} \{\times R_{e}\}_{e \in C} }$ by contracting all the decomposed tensors in $G_{sub,1}$. Similarly, an order-$(L+N-k)$ tensor, $\mathcal{W} \in \mathbb{R}^{ \{\times R_{e}\}_{e \in C} \times I_{p_{k+1}} \times I_{p_{k+2}} \times \cdots \times I_{p_N}}$, can be obtained by contracting all the decomposed tensors in $G_{sub,2}$.

As shown in Figure 5 (Top) in the main text, since the cut $C$ partitions the tensor network $G$ into $G_{sub,1}$ and $G_{sub,2}$, we can write
\begin{equation}
\overrightarrow{\mathcal{X}^{\mathbf{p}}} = \mathcal{Z} \times_{\{C \}} \mathcal{W}
\label{X=ZW}
\end{equation}
where $\mathbf{p} = (p_1, p_2, \ldots, p_N)$ is a mode permutation vector $(1,2,\ldots,N)$.

Equation (\ref{X=ZW}) can be written in an order-$2$ (matrix) format as
\begin{equation}
\mathbf{X}_{[\mathbf{p}, k]} = \mathbf{Z}_{[1:k+L; k]} \mathbf{W}_{[1:L+N-k; L]}
\label{x=zw_matrix}
\end{equation}
where $ \mathbf{Z}_{[1:k+L; k]}  \in \mathbb{R}^{\prod_{n=1}^k I_{p_n} \times \prod_{e \in C} R_e }$ denotes the generalized unfolded version of $\mathcal{Z}$, and  $\mathbf{W}_{[1:L+N-k; L]} \in \mathbb{R}^{\prod_{e \in C} R_e \times \prod_{n=k+1}^N I_{p_n}}$ denotes the unfolded version of $\mathcal{W}$.

Now, as desired, we arrive at the result stated in Observation 1 in the main text
\begin{equation}
\begin{split}
\operatorname{rank}(\mathbf{X}_{[\mathbf{p}, k]}) & \leq \operatorname{min} \left \{\operatorname{rank}(\mathbf{Z}_{[1:k+L; k]}), \operatorname{rank}(\mathbf{W}_{[1:L+N-k;L]}) \right\} \\
& \leq \operatorname{min} \left \{ \prod_{n=1}^k I_{p_n}, \prod_{e \in C} R_e, \prod_{n=k+1}^N I_{p_n}\right\} \leq \prod_{e \in C} R_e
\end{split}
\end{equation}

\section*{Comparing Principal Component Analysis and Canonical Polyadic Decomposition}

\begin{figure*}[h]
    \centering
    \centerline{\includegraphics[width=1\columnwidth]{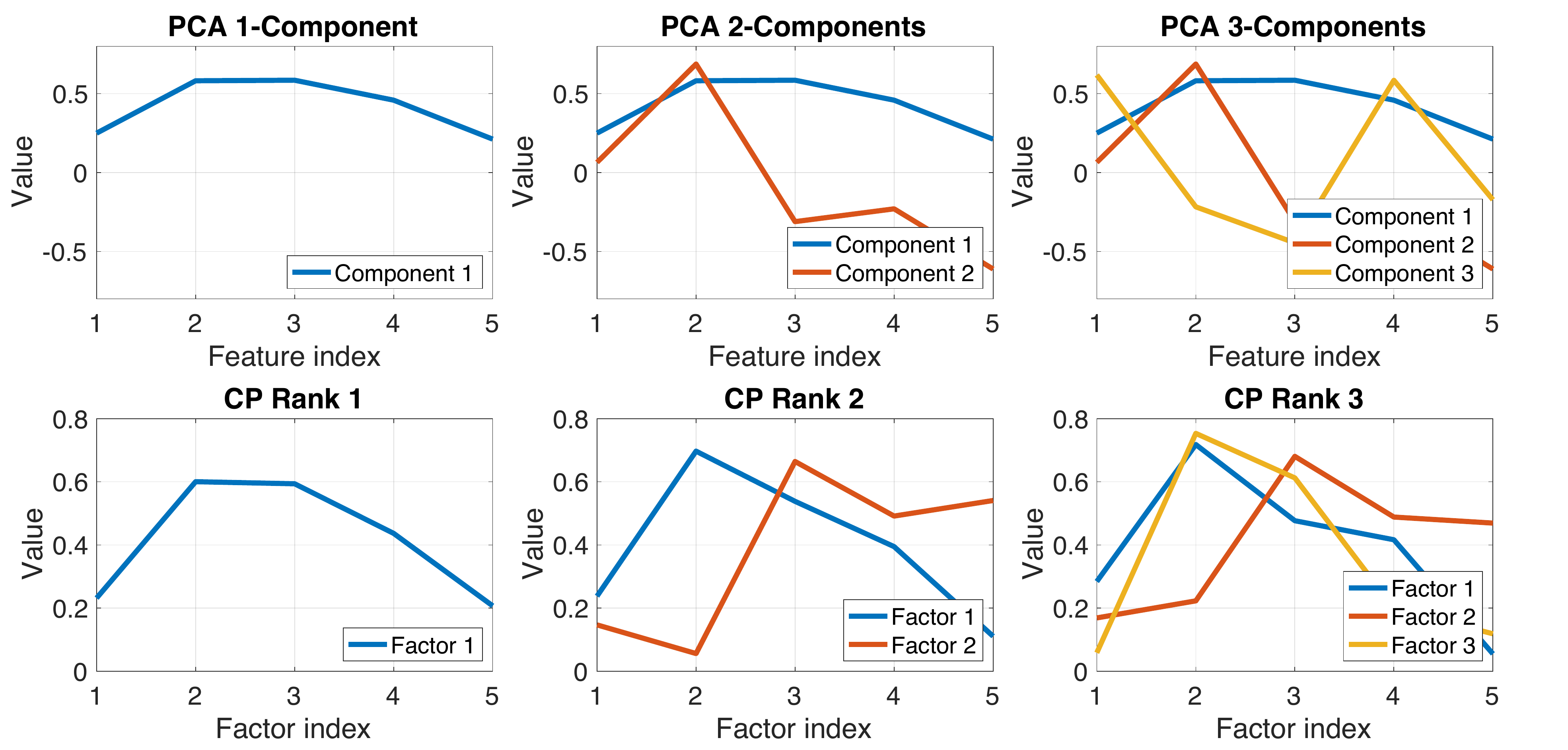}}
    \caption{Behavior of matrix PCA and tensor CP decompositions with varying number of loadings and ranks. \textbf{Top Row:} PCA Loadings of an unfolded random order-$3$ tensor for $1$, $2$, and $3$-component PCA. The first loading always remains the same, independent of the number of extracted components. \textbf{Bottom Row:} Entries of the first factor matrix for rank-$1$, $2$, and $3$ CP decomposition. The first (and second) component(s) change every time when the rank changes.}
    \label{fig:PCA}
\end{figure*}

Unlike matrix methods rooted in linear algebra, the individual components in the ``core'' tensors extracted by tensor network decompositions often change if the TN rank is varied. For example, recall that in the matrix-based PCA, the first principal component always remains the same independent of whether we extract one principal component or many. In contrast, tensor decomposition components are often computed jointly. As a consequence, changing the rank for a CP decomposition algorithm from $R$ to $R+1$ will typically change all of the $R$ previously computed rank-1 components. Figure \ref{fig:PCA} illustrates the difference in the CP components of the first factor matrix when CP decompositions of rank-$1$, $2$, and $3$ are performed on a random order-$3$ tensor. We see that the first CP component changes every time as the CP rank changes, whereas the first loading of PCA remains the same when the number of principal components increases. 

\end{document}